\documentclass[runningheads]{llncs}

\usepackage{eccv}

\usepackage{eccvabbrv}

\usepackage{multirow} 
\usepackage{pifont}
\usepackage{colortbl}
\usepackage{float}
\usepackage{xcolor}
\usepackage{booktabs}
\usepackage{graphicx}
\usepackage{algorithm}
\usepackage{algorithmic}
\usepackage{marvosym}

\usepackage[accsupp]{axessibility}

\usepackage{hyperref}

\usepackage{orcidlink}

\usepackage[most]{tcolorbox}
\tcbset{
  aibox/.style={
    top=10pt,
    colback=blue!2,
    colframe=blue!40!black!70,
    fontupper=\small,
    colbacktitle=blue!30,
    coltitle=black,
    fonttitle=\bfseries,
    enhanced,
    center,
    breakable,
    attach boxed title to top left={yshift=-0.1in,xshift=0.15in},
    boxed title style={boxrule=0pt,colframe=white,},
    fontupper=\footnotesize,
  }
}
\newtcolorbox{AIbox}[2][]{aibox, title=#2,#1}

\begin{document}

\title{Advancing WordArt-Oriented Scene Text Recognition: Datasets and Methods} 

\titlerunning{Advancing WordArt-Oriented Scene Text Recognition}

\author{Xingsong Ye\inst{1,2} \and
Yongkun Du\inst{1,2} \and
Jiaxin Zhang\inst{3} \and Haojie Zhang\inst{4} \and Chong Sun\inst{3} \and
Chen Li\inst{3} \and
Jing Lyu\inst{3} \and
Zhineng Chen\inst{1,2}\textsuperscript{(\Letter)}}

\authorrunning{X.~Ye et al.}

\institute{Institute of Trustworthy Embodied AI, Fudan University \and
Shanghai Key Laboratory of Multimodal Embodied AI, Fudan University\and
WeChat Vision, Tencent Inc. \and South China University of Technology }

\maketitle

\def\thefootnote{\Letter}\footnotetext{Corresponding author. E-mail: zhinchen@fudan.edu.cn}\def\thefootnote{\arabic{footnote}}

\begin{abstract}
WordArt (artistic text) features highly customized fonts, textures, and layouts, making WordArt-oriented scene TExt Recognition (WATER) substantially more challenging than general Scene Text Recognition (STR). Existing STR datasets and methods, typically built around regular scene text and fixed-template inputs, struggle to scale to WATER. Thus, we aim to advance this task from both data and model perspectives.
On the data side, we construct a 2M synthetic dataset, \textbf{WATER-S}, with the scale improved by hundreds of times compared to existing artistic text data. WATER-S consists of two complementary subsets. One rendered by an upgraded rendering pipeline (SynthWordArt), which provides highly accurate and controllable synthetic WordArt data. The other is generated by combining Qwen3-VL for prompt mining and Z-Image for image synthesis, which improves the coverage of realistic and diverse data.
On the model side, we propose \textbf{WATERec}. It adopts a visual encoder supporting arbitrary-shaped inputs and an autoregressive decoder to model complex layouts, structurally breaking the bottleneck of fixed-template STR on WordArt. Experiments show that this architecture outperforms prior STR methods, achieving state-of-the-art performance on irregular texts such as WordArt.
Together with \textbf{WATER-R}, carefully reorganized from existing real STR data, our strong baseline with the new synthetic data and model design reaches 90.40\% accuracy on WordArt-Bench, surpassing both general-purpose and OCR-specialized vision-language models by a large margin. Code and data are available at \url{https://github.com/YesianRohn/WATER}.

\end{abstract}

\section{Introduction}
\label{sec:intro}

\begin{figure}[ht]
 \centering
 \includegraphics[width=0.8\linewidth]{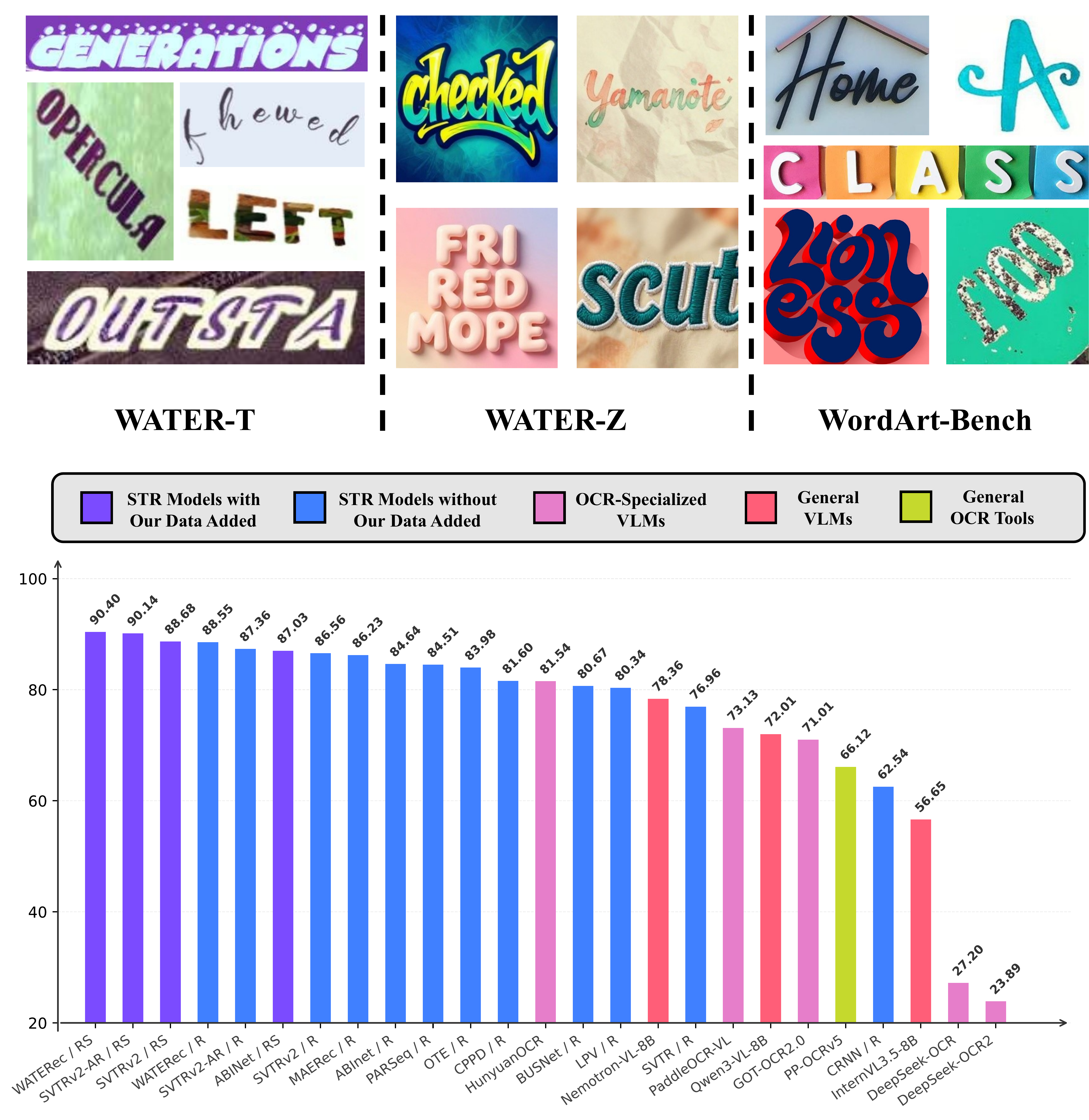}
 \caption{\textbf{Top:} Two subsets' examples of our synthetic datasets (WATER-S) and the real artistic text benchmark (WordArt-Bench). \textbf{Bottom:} Recognition accuracy on WordArt-Bench for various STR methods and VLMs. For each STR entry, the term before ``/'' denotes the model and the term after ``/'' denotes the training data used. ``R'' indicates existing real datasets only, while ``RS'' indicates adding our synthetic data.}
 \label{fig:overview}
\end{figure}

WordArt refers to artistic text~\cite{xie2022toward} that exhibits strong visual stylization and appears widely in posters, signboards, magazines, and other advertising design scenarios. Compared to regular scene text, artistic text is highly customized in font shape, texture, layout, and pattern fusion, as shown in the top of Fig.~\ref{fig:overview}. Designers often embed complex graphics and semantic elements inside character contours, so that the text serves as both a linguistic carrier and a visual symbol. This highly stylized design introduces substantial visual distractions unrelated to the underlying text, making WordArt-oriented scene TExt Recognition (WATER) more challenging than general Scene Text Recognition (STR). Consequently, WATER is often treated as a distinct subtask within STR, and existing work~\cite{jiang2023revisiting, xie2022toward} typically includes dedicated test subsets and benchmarks for this setting. Although mainstream STR models~\cite{abinet, du2025igtr, du2024svtrv2} and recent general/OCR-specialized vision-language models (VLMs)~\cite{cui2025paddleocrvl, team2025hunyuanocr, Qwen3-VL} perform well on common STR benchmarks, their performance on artistic text remains far from satisfactory (see the bottom of Fig.~\ref{fig:overview}).

The primary bottleneck for WATER is the lack of data. 
Real artistic text is difficult to collect in real scenes, and its annotation is expensive and inconsistent. Annotators must carefully inspect characters under complex textures and often must reason about ambiguous shapes to determine the correct label. As a result, current training sets are (i) too small to optimize modern models (e.g., WordArt provides only 4,805 training images~\cite{xie2022toward}), and (ii) insufficiently diverse to cover the long-tail of styles seen in real designs. Breaking this bottleneck requires moving beyond purely manual collection, and large-scale synthesis becomes an almost inevitable choice. 

However, current methods for scene text data generation~\cite{mj, st, textssr} mainly focus on regular text and do not suit WordArt. Synthesizing artistic text is challenging for it must cover broad stylistic variations (fonts, textures, shadow) and realistic effects while still ensuring accurate labels. Therefore, we systematically explore two synthetic data paradigms oriented to WordArt: an improved tool-based rendering approach and a pioneering generative-model-based synthesis pipeline. They emphasize different aspects and, when combined, provide a richer and more comprehensive synthetic dataset \textbf{WATER-S}. Here, we customize and evaluate both paradigms for artistic text.
(1) We collect about 11K open-source artistic fonts, we then develop an artistic-text-oriented rendering engine, \textbf{SynthWordArt}. It pastes target text with specified fonts and layout rules onto background images.  Based on this tool, we generate a 1M-scale synthetic artistic text dataset, \textbf{WATER-T}, which substantially enlarges the number of training samples and the diversity of fonts.
(2) To better mimic the rich style semantics and content compositions in real design scenarios, we leverage the advanced VLM, Qwen3-VL-8B~\cite{Qwen3-VL}, to automatically mine and generate fine-grained captions from existing artistic text images, and obtain about 273K high-quality textual prompts. Then, we employ the state-of-the-art (SOTA) open-source image generation model Z-Image-Turbo~\cite{cai2025z} to synthesize images containing artistic-style text based on these prompts. Finally, we construct another 1M-scale synthetic artistic text dataset, \textbf{WATER-Z}. Compared with traditional rendering, WATER-Z exhibits higher diversity and realism in background textures, layout composition, and global visual style, as depicted at the top of Fig.~\ref{fig:overview}.

Beyond the data, we aim to further establish a strong baseline specifically for WordArt. We find that WordArt often contains highly irregular text with unfixed aspect ratios. But many STR models resize all inputs to a fixed shape, which can severely distort highly irregular text. SVTRv2~\cite{du2024svtrv2} shows that such resizing can dominate failure cases and proposes multi-shape resizing with predefined templates. However, this approach requires manual template design and lacks flexibility in WordArt. Motivated by modern VLMs, we introduce a NaViT-like encoder~\cite{navit} with RoPE~\cite{su2024roformer, heo2024rotary} to support arbitrary-shaped inputs without relying on predefined resizing templates. On the decoding side, WordArt often violates standard left-to-right reading order and requires stronger contextual reasoning. Thus we adopt an autoregressive (AR) decoder to generate the character sequence step by step. Combining these ideas, we present \textbf{WATERec}, a strong STR baseline that better fits the characteristics of the WordArt data.

To fairly and comprehensively evaluate STR methods on artistic text, we also re-construct \textbf{WATER-R} from existing real scene text data (Union14M-L~\cite{jiang2023revisiting}, WordArt-Train~\cite{xie2022toward}, WAS-R~\cite{xie2024dataset}). We perform strict hashing deduplication, so as to avoid label leakage between training and evaluation. Experimental results show that WATERec achieves new SOTA performance on the WordArt-Bench~\cite{xie2022toward} and on several highly challenging subsets (including artistic subset) of the Union14M-Benchmark~\cite{jiang2023revisiting}. When we augment training with our synthetic data, adding either WATER-T or WATER-Z alone already brings consistent and significant improvements. When combining both datasets during training, WATERec further improves its accuracy on the WordArt-Bench to 90.40\%, which, to our knowledge, is the first result exceeding 90\% on this benchmark. We also evaluate representative general VLMs~\cite{Qwen3-VL, wang2025internvl3, deshmukh2025nvidia} and OCR-specialized VLMs~\cite{wei2024general, cui2025paddleocrvl, team2025hunyuanocr, wei2025deepseek, wei2026deepseek} on the WordArt-Bench. Their best accuracies are 78.36\% and 81.54\%, respectively, which remain notably lower than that of our expert baseline. This gap highlights the necessity of dedicated architectural and data design for artistic text.

In summary, our main contributions are:

\begin{itemize}
    \item We build a large-scale synthetic artistic text dataset, WATER-S, consisting of the tool-rendered subset and the model-generated subset. This dataset provides complementary diversity in font style, layout, and visual texture, and offers a scalable data foundation for future research.
    \item We design WATERec to support arbitrary-shaped inputs in STR for the first time and to better model artistic text with complex layouts.
    \item We systematically evaluate representative methods on various STR benchmarks. Combined with WATER-R, WATER-S and WATERec, our strong baseline achieves new SOTA results on WordArt and other challenging scenes.
\end{itemize}

\section{Related Work}

\subsection{Model Perspective on WATER}

From a more general STR perspective, mainstream model architectures can be roughly divided by decoding strategy into Connectionist Temporal Classification~\cite{CTC} (CTC)-based, Parallel Decoding (PD)-based, and AR-based models.
CTC-based methods~\cite{shi2017crnn, zhai2016chinese, duijcai2022svtr, du2024svtrv2} convert 2D images into 1D character sequences via CTC alignment, and work well for regular, horizontally aligned text. Recent variants like SVTRv2~\cite{du2024svtrv2} improve robustness to irregular text by using multi-scale resizing and CTC feature rearrangement.
PD-based methods~\cite{yu2020towards, qiao2021pimnet, lpv, busnet, cppd, du2026mdiff4str} can be viewed as NAR models built on attention mechanisms, predicting all characters in parallel, offering high efficiency but struggling with complex artistic layouts where reading order is ambiguous.
AR-based methods~\cite{Sheng2019nrtr, li2019show, parseq, li2023trocr, ccd, ccdplus, smtr} decode characters step by step (typically left to right), better handling artistic or heavily distorted text through iterative context refinement. CornerTransformer~\cite{xie2022toward} adopts this AR paradigm and is the first baseline tailored to artistic text recognition.
On the general-purpose encoder side, vision backbones have progressed from ViT~\cite{dosovitskiy2020image} to NaViT~\cite{navit} to support arbitrary aspect ratios, often combined with RoPE~\cite{su2024roformer, heo2024rotary} and widely used in modern VLMs~\cite{Qwen3-VL, cui2025paddleocrvl, team2025hunyuanocr}. Inspired by this, we use it to support arbitrary-shaped input, paired with AR decoding as our baseline.

\subsection{Data Perspective on WATER}

Early STR datasets (IIIT5K~\cite{IIIT5K}, ICDAR13~\cite{icdar2013}, SVT~\cite{Wang2011SVT}) focus on regular text in simple conditions. Later benchmarks add irregular text: blurred (ICDAR15~\cite{icdar2015}), perspective-distorted (SVTP~\cite{SVTP}), and curved (CUTE80~\cite{Risnumawan2014cute}) to better match real-world cases.
WordArt~\cite{xie2022toward} is the first dataset and competition~\cite{xie2024icdar} dedicated to artistic text images, derived from TextSeg~\cite{xu2021rethinking}. WAS-R~\cite{xie2024dataset} extends this with more real artistic text. Union14M~\cite{jiang2023revisiting} revisits STR from a data perspective: it collects a large real-world training set and builds a challenging benchmark of seven scenarios, with artistic text as one subset.
Besides real images, synthetic data~\cite{mj, st, yim2021synthtiger, ye2026wrong} plays a crucial role in STR. Earlier methods use tool engines to render mainly regular text, without focusing on artistic styles. Newer work~\cite{textssr, scenevtg} uses diffusion models to generate more diverse and realistic STR data. Here, we follow both directions, combining traditional rendering and diffusion-based generation to build our synthetic WordArt data.

\section{Synthetic Dataset: WATER-S}

Due to the scarcity of real-world artistic text data, training models solely on existing real datasets already reaches a performance bottleneck. However, collecting additional large-scale, high-quality artistic text in the wild is costly and impractical. This limitation makes synthetic data a key component for scaling up training. We therefore design two synthetic data suites that explicitly target two core objectives: controllability and diversity. Along two complementary paths, (1) tool-based rendering and (2) model-based generating, we construct two large-scale synthetic datasets, \textbf{WATER-T} and \textbf{WATER-Z}, collectively referred to as \textbf{WATER-S}. In this section, we describe the construction pipeline, design choices, and statistical properties of WATER-S.

\begin{figure}[t]
 \centering
 \includegraphics[width=0.9\linewidth]{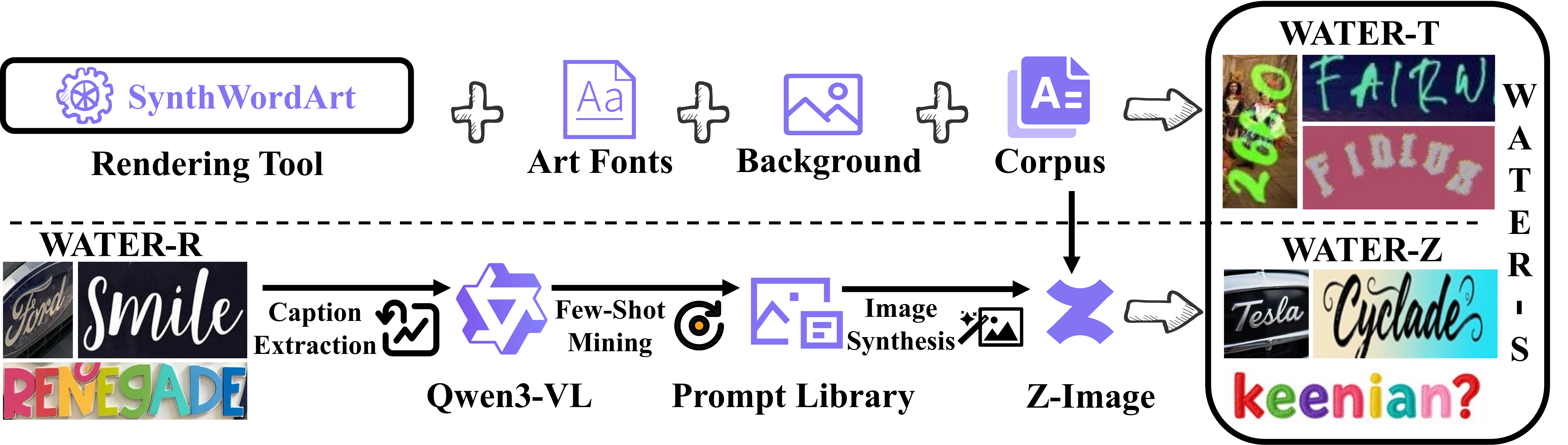}
 \caption{Pipelines of our two synthetic datasets. The top illustrates the WATER-T synthesis pipeline: the art-text-oriented tool SynthWordArt renders provided artistic fonts, background images, and real text corpora into artistic text images. The bottom shows the WATER-Z synthesis pipeline: A small set of real artistic text images are fed into Qwen3-VL (8B)~\cite{Qwen3-VL} to obtain detailed captions. Then Qwen3-VL is leveraged in a few-shot manner to expand these captions into a Prompt Library. Finally Z-Image (Turbo)~\cite{cai2025z} use these prompts to generate more artistic text images.}
 \label{fig:data_pipeline}
\end{figure}

\subsection{Tool-based Rendering: WATER-T}
\label{subsec:water_t}

Building on SynthText~\cite{st} and SynthTIGER~\cite{yim2021synthtiger}, we design \textbf{SynthWordArt}, a rendering engine tailored to artistic text. Our goal is to preserve the precise controllability of classical rendering pipelines over text content and layout, while more faithfully mimicking the diverse layouts and visual distractions that appear in real design scenarios.

Compared with the original synthesis tools, SynthWordArt introduces the following key modifications:
(1) Artistic font support. We replace the commonly used Google Fonts with our library of 11,250 artistic fonts, covering a wide spectrum of artistic styles.
(2) Enhanced layout and typography. Beyond the standard horizontal layout, we add several layout patterns that frequently appear in artistic design, including curved lines, vertical text, multi-orientation layouts, and geometric transformations such as perspective and stretching.

The rest of the pipeline (as shown in the top of Fig.~\ref{fig:data_pipeline}) follows standard rendering tools. For each sample, we randomly sample a text string, a font, a layout configuration, and a background image, and paste the rendered text onto the background. The layout configuration is drawn with explicit sampling ratios, approximately $0.2$ for curved text, $0.3$ for multi-oriented text, and the remainder for general (near-horizontal) layouts. Randomization over these factors yields rich diversity in visual appearance and layout. Using SynthWordArt, we generate 1M synthetic images, denoted as \textbf{WATER-T}.

\subsection{Model-based Generating: WATER-Z}
\label{subsec:water_z}

While tool-based rendering offers strong controllability and perfect text correctness, it still lags behind real designs in terms of global stylistic coherence, semantic texture integration, and overall visual naturalness. To further enhance diversity and realism, we attempt to use the mainstream generative model to synthesize data.

The effectiveness of image generation heavily depends on prompt quality. Simple, template-based prompts are insufficient to cover the rich visual semantics of real design. We therefore propose an automatic few-shot mining pipeline (see Fig.~\ref{fig:data_pipeline}) for artistic text generation.
 Starting from 31,335 samples in the WordArt~\cite{xie2022toward} and WAS-R~\cite{xie2024dataset} training set, we use a VLM (e.g., Qwen3-VL-8B~\cite{Qwen3-VL}) to analyze each artistic text image and generate a detailed caption (please refer to the appendix for the prompt and caption examples). The caption additionally uses an editable placeholder (e.g., \texttt{\textless Text\textgreater}) for the specific word content.
Given these captions, we randomly sample few-shot (e.g., 3) examples and ask the VLM to imitate them and produce a new prompt.
We then apply strict filtering and deduplication to ensure that each prompt contains an editable placeholder and to remove near-duplicate prompts. This pipeline yields 273,488 high-quality prompts tailored for artistic text generation. The prompt resource is independent of any particular generation model, so more advanced models (such as Nano-Banana2) can leverage it to produce even better results.

Here, we adopt Z-Image-Turbo~\cite{cai2025z}, an open-source generative model, as our backbone generator. We use the official default configuration and set the output resolution to $256 \times 256$ for balancing visual fidelity and generation efficiency. Using the prompts described above and randomly substituting the text placeholder with target strings, we synthesize 1M images to construct \textbf{WATER-Z}.

\subsection{Analysis of WATER-T and WATER-Z}
\label{subsec:water_analysis}

Although WATER-T and WATER-Z arise from very different generation mechanisms, they complement each other along several important dimensions:
(1) Font control and text accuracy. WATER-T renders text directly with explicit artistic fonts, so the character shapes and text content are accurate and exactly controllable. In contrast, WATER-Z excels at generating complex materials and globally coherent styles, but its character shapes and legibility are less predictable.
(2) Layout controllability and semantic naturalness. WATER-T allows explicit control over layout patterns and geometric transformations, which enables systematic analysis of how different layouts affect recognition performance. WATER-Z, on the other hand, produces layouts that more closely resemble real designer creations, with more natural composition and stronger semantic integration between foreground text and background.
(3) Distribution coverage. Jointly using WATER-T and WATER-Z for training covers both the ``strongly controlled'' and ``style-diverse'' regimes. This broad coverage is crucial for improving generalization to real artistic text in the wild. It is worth noting that, although the current versions of both datasets focus on English WordArt, the entire generation pipeline is language-agnostic and can be directly adapted to multilingual scenarios by replacing the underlying text corpus.

\section{Strong Baseline: WATERec}

\subsection{Overall Architecture}

\begin{figure}[ht]
 \centering
 \includegraphics[width=0.95\linewidth]{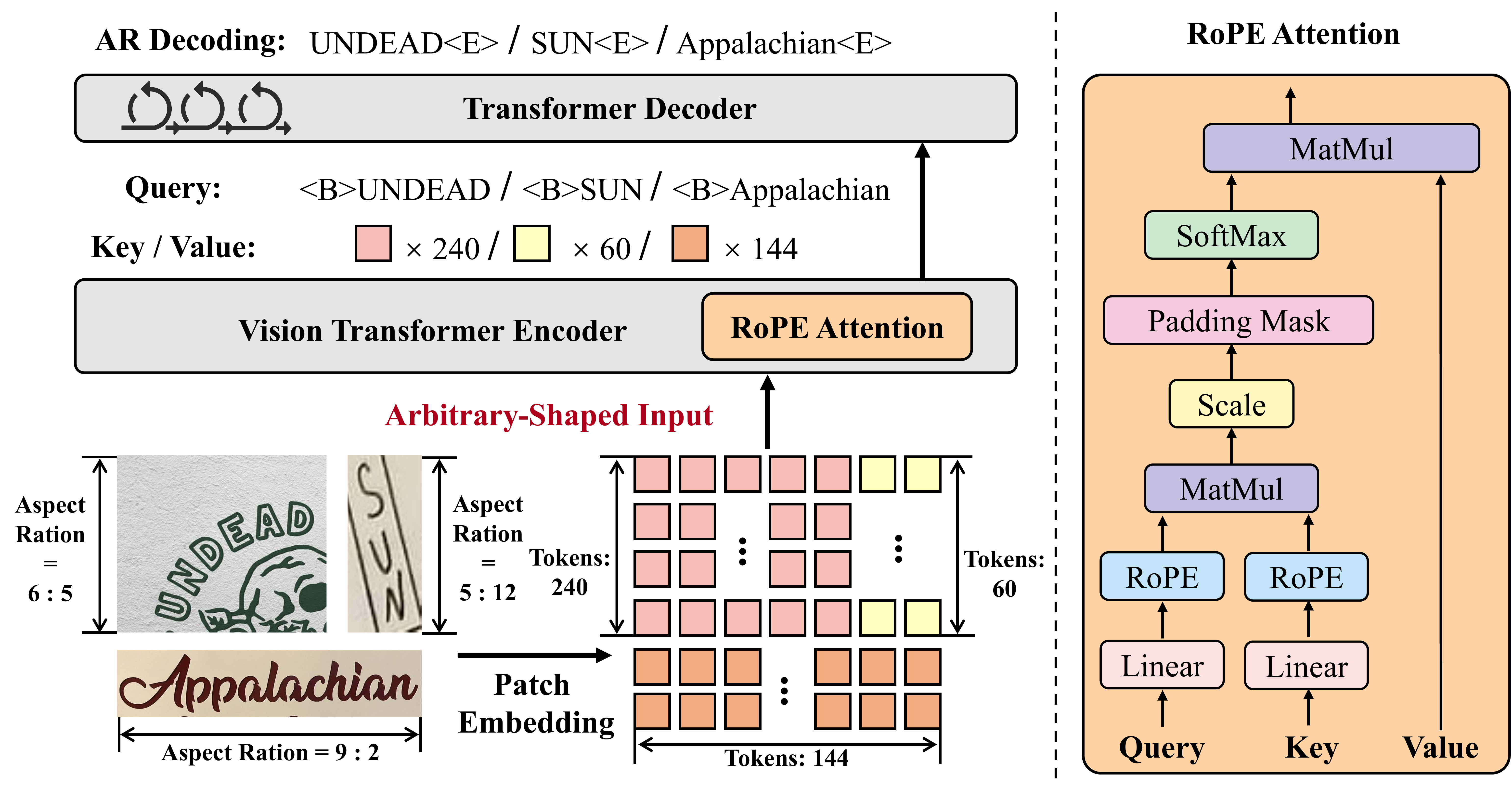}
 \caption{The overall architecture of WATERec. It consists of a Vision Transformer Encoder that uses RoPE Attention (illustrated on the right) to support inputs of arbitrary shape, and an AR Transformer Decoder. It also shows images of different sizes, which are processed into tokens of different lengths. \texttt{<B>} denotes the beginning token of decoding, and \texttt{<E>} denotes the end token of decoding.}
 \label{fig:model}
\end{figure}

WATERec aims to provide a simple yet effective baseline for artistic text recognition. The model (as shown in Fig.~\ref{fig:model}) is fully generic: it does not rely on any fixed-shape template, it handles images with arbitrary aspect ratios, and it models highly unconstrained layouts and visual styles in a unified way.

As illustrated in Fig.~\ref{fig:model}, WATERec adopts a simple encoder design that accepts inputs of arbitrary shape, followed by an AR Transformer decoder. The encoder is a 6-layer Transformer with RoPE attention~\cite{su2024roformer, heo2024rotary} that processes visual token sequences of varying length (see details below), and the decoder uses 2 cross-attention layers to attend to the encoder outputs and predicts characters one by one under a standard cross-entropy loss. This design improves robustness to non-standard reading orders, curved or circular layouts, vertical text, and other complex artistic typesetting styles.

\subsection{Arbitrary-Shaped Input}

Most traditional STR methods~\cite{shi2017crnn, abinet, jiang2023revisiting} assume fixed-template inputs (e.g., $32\times128$), or rely on some predefined templates~\cite{du2024svtrv2}. These designs work well for general horizontal text in natural scenes, but they show clear limitations in artistic text scenarios.  Artistic text exhibits extreme variation in aspect ratio, ranging from very long horizontal titles to almost square or vertical text. 

Therefore, WATERec adopts a visual encoding strategy that naturally adapts to arbitrary aspect ratios. Concretely, given an input image $\mathbf{I}\in\mathbb{R}^{H\times W\times 3}$ of arbitrary resolution, we first rescale it while preserving the aspect ratio to $\hat{\mathbf{I}}\in\mathbb{R}^{\hat{H}\times\hat{W}\times 3}$. The resulting number of tokens $N=(\hat{H}/p)\times(\hat{W}/p)$ lies in a predefined range $[64, 256]$, where $p=4$ is the patch size. The upper bound ensures a fair comparison with existing methods and the lower bound prevents very small images from being represented with too few tokens. We then linearly project each image patch into a $d$-dimensional visual token ($d = 384$ to align with prior work), yielding a token sequence $\mathbf{X}\in\mathbb{R}^{N\times d}$. We arrange these patch tokens in a row-major order to obtain a 1D sequence. The WATERec encoder maps $\mathbf{X}$ to contextual features $\mathbf{Z}\in\mathbb{R}^{N\times d}$, which are then consumed by the AR decoder through cross-attention to autoregressively produce the character sequence $\{y_1,\dots,y_T\}$ ($T\le25$). To support efficient batch training and robust inference, we apply padding to sequences of different lengths at the batch level and use the corresponding padding masks in attention computation.

A remaining challenge is how the encoder above should perceive the spatial layout of arbitrary-shaped inputs, which is crucial for recognizing text under complex artistic typesetting. Absolute and sinusoidal position embeddings (APE and SPE) are highly sensitive to image size and shape, and training data cannot cover all resolutions and layouts. This motivates our choice of the RoPE design~\cite{heo2024rotary} for the visual encoder of WATERec. The core idea is to apply position-dependent rotations to the query (Q) and key (K) vectors, so that the dot-product attention implicitly encodes relative positional information and generalizes naturally across the variable token sequence length $N$. For more details of the three position embeddings, please refer to the appendix.

\section{Experiments}

\subsection{Datasets and Implementation Details}

\subsubsection{Training data.}
In addition to the synthetic datasets introduced earlier, we construct a large-scale real-world training set by combining three sources: the widely used Union14M-L~\cite{jiang2023revisiting} for STR, the training split of WordArt~\cite{xie2022toward}, and the WAS-R~\cite{xie2024dataset}.  
To avoid label leakage caused by overlapping images between training and evaluation sets, we perform strict hashing deduplication between the merged real dataset and all test sets used in our experiments.  
The resulting real training set is denoted as \textbf{WATER-R} and contains 3,225,130 text instances.

\subsubsection{Test data.}
For artistic text evaluation, we use the test split of WordArt~\cite{xie2022toward} with 1,511 images and refer to this benchmark as \textbf{A-Bench}.  
To assess the generalization of WATERec, we also evaluate on six common benchmarks (collectively denoted as \textbf{C-Bench}): ICDAR 2013 (IC13)~\cite{icdar2013}, Street View Text (SVT)~\cite{Wang2011SVT}, IIIT5K-Words (IIIT5K)~\cite{IIIT5K}, ICDAR 2015 (IC15)~\cite{icdar2015}, Street View Text-Perspective (SVTP)~\cite{SVTP}, and CUTE80 (CUTE)~\cite{Risnumawan2014cute}.  
In addition, we report results on the Union14M-Benchmark (\textbf{U-Bench})~\cite{jiang2023revisiting}, with seven subsets: Curve (\textit{CUR}), Multi-Oriented (\textit{MLO}), \textbf{Artistic} (\textit{ART}), Contextless (\textit{CTL}), Salient (\textit{SAL}), Multi-Words (\textit{MLW}), and General (\textit{GEN}).

\subsubsection{Training setup.}
We train WATERec and all comparison models (with their official configurations unchanged) using the OpenOCR framework.  
For WATERec, we largely follow the training protocol of SVTRv2~\cite{du2024svtrv2}. We use the AdamW optimizer~\cite{adamw} with a weight decay of 0.05, a base learning rate of $6.5\times10^{-4}$, and a total batch size of 2048.  
We adopt a one-cycle learning rate scheduler with 1.5 epochs of linear warm-up over 20 training epochs.  
Data augmentation follows PARSeq~\cite{parseq} and includes random rotation, perspective distortion, motion blur, and Gaussian noise.  
The maximum text length during training is 25.  
The character set contains 94 tokens, including digits, letters, and common symbols.  
All models are trained on 8 NVIDIA V100 GPUs.

\begin{table*}[ht]
  \centering
  \resizebox{\textwidth}{!}{%
  \setlength\tabcolsep{1pt}
\begin{tabular}{c|c|c|ccccccc|cccccccc}
\toprule
\multirow{2}{*}{\textbf{Type}} &\multirow{2}{*}{\textbf{Model}} & \textbf{WordArt} & \multicolumn{7}{c|}{\textbf{Common Benchmarks}}                                                                & \multicolumn{8}{c}{\textbf{Union14M-Benchmark}}                                             \\
 & & \textbf{Benchmark} &
\textit{IIIT} & \textit{SVT}  & \textit{IC13} & \textit{IC15} & \textit{SVTP} & \textit{CUTE} & \textit{AVG} &
\textit{CUR} & \textit{MLO} & \textit{ART} & \textit{CTL} & \textit{SAL} & \textit{MLW} & \textit{GEN} & \textit{AVG}                  
\\
\midrule

\multirow{3}{*}{\textbf{CTC}} 
& CRNN~\cite{shi2017crnn}    
& 62.54 & 93.07 & 87.02 & 92.30 & 80.95 & 80.47 & 82.64 & 86.07 & 27.21 & 35.65 & 39.00 & 53.15 & 23.69 & 48.18 & 64.53 & 41.63 \\
& SVTR~\cite{duijcai2022svtr}    
& 76.96 & 97.30 & 94.74 & 96.38 & 87.47 & 89.30 & 93.75 & 93.16 & 68.38 & 75.46 & 62.67 & 71.50 & 66.20 & 71.60 & 77.00 & 70.40 \\
& SVTRv2~\cite{du2024svtrv2}    
& 86.56 & 99.17 & \textbf{98.45} & \textbf{99.18} & 90.23 & 94.73 & \textbf{98.96} & 96.79 & 90.48 & 89.26 & 79.78 & \textbf{86.14} & 85.66 & 86.29 & 85.37 & 86.14 \\
\midrule
\multirow{4}{*}{\textbf{PD}} 
& ABINet~\cite{abinet}    
& 84.64 & 98.73 & 97.37 & 97.67 & 90.01 & 94.73 & 98.26 & 96.13 & 83.76 & 90.14 & 72.00 & 77.54 & 80.29 & 78.64 & 81.71 & 80.58 \\
& LPV~\cite{lpv} 
& 80.34 & 97.93 & 95.05 & 96.62 & 87.69 & 91.47 & 96.18 & 94.16 & 76.17 & 84.59 & 64.44 & 73.94 & 74.42 & 72.69 & 79.37 & 75.09 \\
& BUSNet~\cite{busnet} 
& 80.67 & 98.00 & 95.52 & 97.55 & 88.79 & 93.33 & 95.83 & 94.84 & 79.97 & 87.14 & 66.22 & 74.20 & 76.18 & 68.81 & 81.28 & 76.26 \\
& CPPD~\cite{cppd} 
& 81.60 & 97.00 & 96.29 & 97.43 & 87.02 & 92.71 & 96.18 & 94.44 & 85.20 & 84.30 & 72.00 & 77.92 & 79.66 & 75.12 & 80.99 & 79.31 \\
\midrule
\multirow{5}{*}{\textbf{AR}} 
& PARSeq~\cite{parseq}    
& 84.51 & 98.40 & 97.68 & 98.13 & 89.40 & 95.50 & 97.57 & 96.11 & 84.71 & 92.26 & 74.67 & 80.23 & 80.35 & 79.61 & 84.02 & 82.26 \\
& MAERec~\cite{jiang2023revisiting}  
& 86.23 & 99.13 & 97.99 & 99.07 & 91.05 & \textbf{96.28} & 98.26 & \textbf{96.96} & 90.60 & 93.64 & 78.00 & 84.34 & 86.86 & 86.53 & \textbf{85.65} & 86.52 \\
& OTE~\cite{ote}   
& 83.98 & 97.87 & 96.75 & 97.20 & 89.18 & 93.33 & 96.53 & 95.14 & 81.16 & 88.09 & 72.78 & 73.81 & 76.69 & 62.50 & 81.38 & 76.63 \\
& SVTRv2-AR~\cite{du2024svtrv2}   
& 87.36 & \textbf{99.23} & 97.06 & 98.72 & 90.61 & 95.35 & 97.92 & 96.48 & 91.59 & \textbf{95.25} & 80.33 & 86.01 & 87.49 & 87.14 & 85.58 & 87.63 \\
& WATERec 
& \textbf{88.55} & 99.17 & 97.68 & 98.72 & \textbf{91.11} & 95.19 & 98.26 & 96.69 & \textbf{93.03} & 94.89 & \textbf{80.56} & 85.62 & \textbf{89.39} & \textbf{88.35} & 85.16 & \textbf{88.14} \\

\bottomrule
\end{tabular}
  }
\caption{Quantitative comparison of different STR models trained on the WATER-R dataset. The values are all percentages (\%), and AVG represents the arithmetic mean.}
\label{tab:comp_model}
\end{table*}

\subsection{Experimental Results}

\subsubsection{STR model comparison.}
Tab.~\ref{tab:comp_model} reports a systematic comparison of the three mainstream STR paradigms.  
Among NAR (CTC and PD) methods, SVTRv2~\cite{du2024svtrv2} achieves strong performance on both A-Bench and U-Bench, benefiting from its design for handling multi-scale inputs.  
Building on this architecture, its AR variant SVTRv2-AR further improves accuracy on artistic text, indicating that coupling attention-based sequence modeling with strong visual features is especially beneficial in complex scenes.  
WATERec follows the AR paradigm but preserves the original aspect ratio of artistic text images as much as possible. 
This design yields the best average accuracy on both A-Bench and U-Bench, demonstrating that retaining the native width/height ratio is particularly effective for challenging subsets such as Artistic and Salient.  
However, this comes with a trade-off: on the more regular and simple C-Bench, the fixed-resolution MAERec~\cite{jiang2023revisiting} slightly outperforms SVTRv2 and WATERec.  
This suggests that enforcing a unified input resolution remains advantageous in standard, less distorted scenarios.

\subsubsection{VLM capabilities.}
On A-Bench, we further evaluate mainstream OCR systems, including general VLMs (Qwen3-VL-8B~\cite{Qwen3-VL}, InternVL3.5-8B~\cite{wang2025internvl3}, Nemotron-VL-8B~\cite{deshmukh2025nvidia}) and OCR-specialized VLMs / general OCR tools (GOT-OCR 2.0~\cite{wei2024general}, PaddleOCR-VL~\cite{cui2025paddleocrvl}, HunyuanOCR~\cite{team2025hunyuanocr}, DeepSeek-OCR~\cite{wei2025deepseek}, DeepSeek-OCR2~\cite{wei2026deepseek}, PP-OCRv5~\cite{cui2025paddleocr}).  
As shown in Fig.~\ref{fig:overview}, their accuracies on A-Bench mostly fall in the 70\%--80\% range, with HunyuanOCR performing best at 81.54\%.  
While VLMs already perform strongly in general OCR settings, they still lack robustness and fine-grained perception in artistic text scenarios with severe deformations, diverse styles, and complex layouts.  
This gap highlights artistic text recognition as a challenging and important direction that merits further study.

\begin{table*}[ht]
  \centering
  \resizebox{\textwidth}{!}{%
  \setlength\tabcolsep{1pt}
\begin{tabular}{c|c|c|ccccccc|cccccccc}
\toprule
\multirow{2}{*}{\textbf{Training Data}} & \multirow{2}{*}{\textbf{Volume}} & \textbf{WordArt} & \multicolumn{7}{c|}{\textbf{Common Benchmarks}}                                                                & \multicolumn{8}{c}{\textbf{Union14M-Benchmark}}                                             \\
&  & \textbf{Benchmark} &
\textit{IIIT} & \textit{SVT}  & \textit{IC13} & \textit{IC15} & \textit{SVTP} & \textit{CUTE} & \textit{AVG} &
\textit{CUR} & \textit{MLO} & \textit{ART} & \textit{CTL} & \textit{SAL} & \textit{MLW} & \textit{GEN} & \textit{AVG}                  
\\
\midrule
WATER-R   & 3.2M  & 88.55 & 99.17 & 97.68 & 98.72 & 91.11 & 95.19 & 98.26 & 96.69 & 93.03 & 94.89 & 80.56 & 85.62 & 89.39 & 88.35 & 85.16 & 88.14 \\
\midrule
+ WATER-T  & +0.5M  & 89.54 & 99.33 & 98.30 & 98.83 & 90.72 & 95.35 & 98.61 & 96.86 & 93.98 & 95.69 & 82.44 & 86.26 & \textbf{90.08} & 89.44 & 85.52 & 89.06 \\
+ WATER-T  & +1M    & 89.81 & 99.33 & \textbf{98.61} & \textbf{98.95} & 90.89 & 96.43 & 98.96 & \textbf{97.20} & 94.27 & 96.20 & 83.22 & 87.16 & 90.02 & 89.44 & 85.60 & 89.42 \\
+ WATER-Z  & +0.5M  & 88.91 & 99.07 & 98.30 & \textbf{98.95} & 90.56 & 95.66 & 97.92 & 96.74 & 93.65 & 95.11 & 81.56 & 86.39 & 89.70 & 89.20 & 85.25 & 88.69 \\
+ WATER-Z  & +1M    & 89.41 & 99.17 & \textbf{98.61} & 98.37 & 90.89 & 96.12 & 97.57 & 96.79 & 93.57 & 94.96 & 80.56 & 84.98 & 89.51 & 89.56 & 85.06 & 88.31 \\
+ WATER-S  & +1M    & 89.94 & 99.33 & 98.45 & \textbf{98.95} & \textbf{91.44} & 95.66 & 98.61 & 97.07 & 94.27 & 95.25 & 83.11 & 86.65 & 89.58 & 89.56 & 85.40 & 89.12 \\
+ WATER-S  & +2M    & \textbf{90.40} & \textbf{99.43} & 98.30 & 98.48 & 91.22 & 95.66 & 98.96 & 97.01 & 94.27 & 95.54 & \textbf{84.11} & 86.91 & 89.89 & 89.32 & 85.60 & 89.38 \\
+ WATER-S  & +3M    & 90.07 & 99.17 & 98.15 & 98.83 & 90.83 & 96.12 & 98.96 & 97.01 & \textbf{94.31} & \textbf{96.35} & 83.11 & \textbf{87.55} & 89.83 & 89.32 & \textbf{85.62} & \textbf{89.44} \\
\midrule
+SynthText~\cite{st}       & +1M    & 88.95 & 99.10 & \textbf{98.61} & 98.83 & 91.00 & \textbf{96.59} & 98.61 & 97.12 & 93.94 & 95.76 & 82.56 & 86.39 & 89.58 & \textbf{89.81} & 85.46 & 89.07 \\
+ SynthTIGER~\cite{yim2021synthtiger}& +1M    & 88.55 & 99.23 & 98.30 & 98.83 & 91.28 & 95.35 & \textbf{99.65} & 97.11 & 93.49 & 94.89 & 80.33 & 86.26 & 88.76 & 89.08 & 85.47 & 88.32 \\
+ TextSSR~\cite{textssr}& +1M    & 88.09 & 99.30 & 97.99 & 98.48 & 90.89 & 96.28 & 98.61 & 96.93 & 93.57 & 95.62 & 82.22 & 85.88 & 89.07 & 88.71 & 85.28 & 88.62 \\
\midrule
WATER-S & 2M    & 80.94 & 97.53	& 93.97	& 96.97	& 80.23	& 86.67	& 95.14	& 91.75	& 77.29	& 81.52	& 71.22	& 81.90	& 69.80	& 83.25	& 54.76	& 74.25 \\

\bottomrule
\end{tabular}
  }
\caption{Performance comparison of the gains brought by synthetic data from different sources and at different scales to the real data.}
\label{tab:comp_data}
\end{table*}

\subsubsection{Synthetic data comparison.}
Tab.~\ref{tab:comp_data} analyzes the effect of adding our synthetic artistic data on top of the real dataset WATER-R.  
We observe consistent and significant gains on both artistic and general benchmarks.  
Among different synthetic sources, WATER-S, which combines tool-rendered and model-based generation, shows the most robust performance across scales.  
On A-Bench, adding 2M WATER-S samples to WATER-R improves accuracy by 1.85\%.  
On C-Bench and U-Bench, it also yields additional gains.  
These results indicate that moderately scaled, high-quality, and diverse artistic synthetic data is crucial for improving model performance.  
Comparing different synthetic types, WATER-T  consistently outperforms conventional tool-rendered datasets such as SynthText~\cite{st} and SynthTIGER~\cite{yim2021synthtiger} on both artistic and general benchmarks.  
This shows that explicitly injecting artistic fonts and complex layouts into the traditional synthesis pipeline is both effective and necessary.  
At the same time, WATER-Z outperforms general diffusion-based STR data synthesis methods like TextSSR~\cite{textssr} on artistic text, demonstrating the necessity of WordArt data generation. 
Moreover, WATER-T and WATER-Z are naturally complementary in visual style and scene diversity.  
For example, 1M WATER-S constructed from 0.5M WATER-T and 0.5M WATER-Z outperforms each of their respective 1M counterparts. Training on WATER-S alone still delivers strong performance, further demonstrating the potential of our synthetic data.

\subsection{Ablation Study}

\subsubsection{Scale of synthetic data.}
We first study the marginal gains of WATER-S at different scales, as reported in Tab.~\ref{tab:comp_data}.  
The improvement from synthetic data follows a ``rise-then-saturate'' pattern. 
When increasing WATER-S from 1M to 2M samples, the gains are consistent across A-Bench and most U-Bench subsets. It suggests that at this scale, the additional diversity in style, fonts, and scenes compensates well for the coverage gaps in real data while keeping noise under control.  
However, when further scaling WATER-S to 3M, the overall gains become marginal and some subsets even show slight fluctuations.  
Given that the real data size is fixed, too many synthetic samples introduce distribution mismatch and noise accumulation, which offsets the benefits. 
In our setting, a synthetic-to-real ratio of roughly 2:3 strikes a favorable balance between improving generalization and controlling noise.

\begin{table*}[ht]
  \centering
  \resizebox{\textwidth}{!}{%
  \setlength\tabcolsep{1pt}
\begin{tabular}{c|c|c|ccccccc|cccccccc}
\toprule
\multirow{2}{*}{\textbf{Model}} & \multirow{2}{*}{\textbf{Training Data}} & \textbf{WordArt} & \multicolumn{7}{c|}{\textbf{Common Benchmarks}}                                                                & \multicolumn{8}{c}{\textbf{Union14M-Benchmark}}                                             \\
&  & \textbf{Benchmark} &
\textit{IIIT} & \textit{SVT}  & \textit{IC13} & \textit{IC15} & \textit{SVTP} & \textit{CUTE} & \textit{AVG} &
\textit{CUR} & \textit{MLO} & \textit{ART} & \textit{CTL} & \textit{SAL} & \textit{MLW} & \textit{GEN} & \textit{AVG}                  
\\
\midrule
\multirow{3}{*}{SVTRv2~\cite{du2024svtrv2}}   & WATER-R  & 86.56 & 99.17 & 98.45 & 99.18 & 90.23 & 94.73 & 98.96 & 96.79 & 90.48 & 89.26 & 79.78 & 86.14 & 85.66 & 86.29 & 85.37 & 86.14 \\
  & +WATER-S  & 88.68 & 99.33 & 98.15 & 98.83 & 90.89 & 95.19 & 99.31 & 96.95 & 92.50 & 92.62 & 82.11 & 87.68 & 88.63 & 89.08 & 84.86 & 88.21 \\
& $\Delta$ & 
\textcolor{blue}{+2.12} & 
\textcolor{blue}{+0.17} & 
\textcolor{red}{-0.31} & 
\textcolor{red}{-0.35} & 
\textcolor{blue}{+0.66} & 
\textcolor{blue}{+0.47} & 
\textcolor{blue}{+0.35} & 
\textcolor{blue}{+0.16} & 
\textcolor{blue}{+2.02} & 
\textcolor{blue}{+3.36} & 
\textcolor{blue}{+2.33} & 
\textcolor{blue}{+1.54} & 
\textcolor{blue}{+2.97} & 
\textcolor{blue}{+2.79} & 
\textcolor{red}{-0.51} & 
\textcolor{blue}{+2.07} \\
\midrule
\multirow{3}{*}{ABINet~\cite{abinet}}   & WATER-R  & 84.64 & 98.73 & 97.37 & 97.67 & 90.01 & 94.73 & 98.26 & 96.13 & 83.76 & 90.14 & 72.00 & 77.54 & 80.29 & 78.64 & 81.71 & 80.58 \\
  & +WATER-S  & 87.03 & 98.40 & 97.68 & 98.25 & 90.28 & 94.42 & 97.92 & 96.16 & 86.07 & 91.16 & 77.11 & 80.49 & 83.64 & 83.37 & 81.58 & 83.35 \\
& $\Delta$ & 
\textcolor{blue}{+2.39} & 
\textcolor{red}{-0.33} & 
\textcolor{blue}{+0.31} & 
\textcolor{blue}{+0.58} & 
\textcolor{blue}{+0.28} & 
\textcolor{red}{-0.31} & 
\textcolor{red}{-0.35} & 
\textcolor{blue}{+0.03} & 
\textcolor{blue}{+2.31} & 
\textcolor{blue}{+1.02} & 
\textcolor{blue}{+5.11} & 
\textcolor{blue}{+2.95} & 
\textcolor{blue}{+3.35} & 
\textcolor{blue}{+4.73} & 
\textcolor{red}{-0.13} & 
\textcolor{blue}{+2.76} \\
\midrule
\multirow{3}{*}{SVTRv2-AR}   & WATER-R  & 87.36 & 99.23 & 97.06 & 98.72 & 90.61 & 95.35 & 97.92 & 96.48 & 91.59 & 95.25 & 80.33 & 86.01 & 87.49 & 87.14 & 85.58 & 87.63 \\
  & +WATER-S  & 90.14 & 99.17 & 98.30 & 99.18 & 91.28 & 95.50 & 98.61 & 97.01 & 94.19 & 95.69 & 84.67 & 86.26 & 89.51 & 89.56 & 85.65 & 89.36 \\
& $\Delta$ & 
\textcolor{blue}{+2.78} & 
\textcolor{red}{-0.07} & 
\textcolor{blue}{+1.24} & 
\textcolor{blue}{+0.47} & 
\textcolor{blue}{+0.66} & 
\textcolor{blue}{+0.16} & 
\textcolor{blue}{+0.69} & 
\textcolor{blue}{+0.52} & 
\textcolor{blue}{+2.60} & 
\textcolor{blue}{+0.44} & 
\textcolor{blue}{+4.33} & 
\textcolor{blue}{+0.25} & 
\textcolor{blue}{+2.02} & 
\textcolor{blue}{+2.43} & 
\textcolor{blue}{+0.07} & 
\textcolor{blue}{+1.73} \\
\midrule
\multirow{3}{*}{WATERec}   & WATER-R  & 88.55 & 99.17 & 97.68 & 98.72 & 91.11 & 95.19 & 98.26 & 96.69 & 93.03 & 94.89 & 80.56 & 85.62 & 89.39 & 88.35 & 85.16 & 88.14 \\
  & +WATER-S  & 90.40 & 99.43 & 98.30 & 98.48 & 91.22 & 95.66 & 98.96 & 97.01 & 94.27 & 95.54 & 84.11 & 86.91 & 89.89 & 89.32 & 85.60 & 89.38 \\
& $\Delta$ & 
\textcolor{blue}{+1.85} & 
\textcolor{blue}{+0.27} & 
\textcolor{blue}{+0.62} & 
\textcolor{red}{-0.23} & 
\textcolor{blue}{+0.11} & 
\textcolor{blue}{+0.47} & 
\textcolor{blue}{+0.69} & 
\textcolor{blue}{+0.32} & 
\textcolor{blue}{+1.24} & 
\textcolor{blue}{+0.66} & 
\textcolor{blue}{+3.56} & 
\textcolor{blue}{+1.28} & 
\textcolor{blue}{+0.51} & 
\textcolor{blue}{+0.97} & 
\textcolor{blue}{+0.43} & 
\textcolor{blue}{+1.24} \\
\bottomrule
\end{tabular}
  }
\caption{Effect of adding our WATER-S (2M) synthetic data to different STR models.}
\label{tab:comp_data_add}
\end{table*}

\subsubsection{Generality of synthetic data.}
We then validate the generality and robustness of WATER-S across different STR architectures, as shown in Tab.~\ref{tab:comp_data_add}.  
We consider a CTC-based model (SVTRv2~\cite{du2024svtrv2}), a PD-based model (ABINet~\cite{abinet}), another AR-based model (SVTRv2-AR), and our WATERec.  
For all four models, adding WATER-S on top of the same real dataset WATER-R leads to consistent and significant improvements on A-Bench, with gains ranging from +1.85\% to +2.78\%.  
This cross-architecture benefit indicates that high-quality artistic synthetic data is broadly useful rather than tailored to a specific model design.  
On the more challenging U-Bench, all models also obtain 1\%–4\% improvements on most subsets, with especially notable gains on the Artistic subset.  

\begin{table*}[ht]
  \centering
  \resizebox{\textwidth}{!}{%
  \setlength\tabcolsep{1pt}
  \begin{tabular}{ccc|c|ccc|cccccccc}
    \toprule
   \textbf{Arbitrary} &  \textbf{Position} & \textbf{Token} & \textbf{WordArt} & \multicolumn{3}{c|}{\textbf{Common Benchmarks}} & \multicolumn{8}{c}{\textbf{Union14M-Benchmark}} \\
  \textbf{Shape}  & \textbf{Encoding}  &  \textbf{Range} & \textbf{Benchmark} & Regular & Irregular  & Average & \textit{CUR} & \textit{MLO} & \textit{ART} & \textit{CTL} & \textit{SAL} & \textit{MLW} & \textit{GEN} & \textit{AVG} \\
    \midrule
    \ding{55} & NoPE   & [256, 256] & 86.83 & 98.25 & 95.15 & 96.70 & 90.15 & 94.52 & 77.78 & 84.34 & 86.73 & 87.86 & 85.41 & 86.69 \\
    \ding{51} & NoPE   & [64, 256]  & 49.57 & 89.14 & 86.07 & 87.60 & 60.76 & 61.29 & 47.56 & 54.17 & 52.37 & 47.21 & 74.94 & 56.90 \\
    \ding{51} & APE    & [64, 256]  & 87.69 & 98.29 & 94.41 & 96.35 & 92.00 & 93.86 & 79.00 & 85.24 & 88.63 & 86.89 & 84.08 & 87.10 \\
    \ding{51} & SPE    & [64, 256]  & 87.29 & 98.36 & 94.37 & 96.37 & 92.25 & 94.08 & 79.00 & 84.21 & 88.12 & 86.17 & 84.29 & 86.87 \\
    \ding{55} & RoPE   & [256, 256] & 87.88 & \textbf{98.71} & 94.99 & 96.85 & 92.09 & 90.94 & 80.22 & 85.11 & 87.87 & 87.14 & \textbf{85.55} & 86.99 \\
    \ding{51} & RoPE   & [1, 256]   & 88.29 & 98.47 & 94.80 & 96.64 & 93.36 & 94.96 & 80.11 & \textbf{85.62} & \textbf{89.58} & 89.20 & 78.81 & 87.38 \\
    \ding{51} & RoPE   & [64, 256]  & 88.55 & 98.52 & 94.86 & 96.69 & 93.03 & 94.89 & 80.56 & \textbf{85.62} & 89.39 & 88.35 & 85.16 & 88.14 \\
    \ding{51} & RoPE   & [64, 512]  & \textbf{88.82} & 98.61 & \textbf{95.75} & \textbf{97.18} & \textbf{94.39} & \textbf{95.62} & \textbf{83.22} & 85.49 & 89.32 & \textbf{89.81} & 85.53 & \textbf{89.06} \\
    \bottomrule
\end{tabular}
  }
\caption{Ablation study on model design choices, including arbitrary shape modeling, different position encoding schemes, and token range settings. ``Regular'' denotes the average on \textit{IIIT}, \textit{SVT}, and \textit{IC13} datasets,``Irregular'' represents the average on \textit{IC15}, \textit{SVTP}, and \textit{CUTE} datasets. Training settings are consistent with those in Tab.~\ref{tab:comp_model}.}
\label{abl}
\end{table*}

\subsubsection{Model components.}
To understand the contribution of each component in WATERec, we perform ablations on arbitrary-shape modeling, positional encoding, and token range, as summarized in Tab.~\ref{abl}.  
Under arbitrary-shape modeling, adding a suitable positional encoding brings substantial improvements, and RoPE offers greater advantages compared with APE and SPE.
Removing positional encoding (NoPE) entirely under arbitrary shapes leads to a large performance drop.  
This indicates that arbitrary-shape modeling provides the structural capacity to capture complex geometric deformations in artistic text, but its benefit is only fully realized when combined with an appropriate sequence modeling strategy.  
Expanding the range from a fixed $[256,256]$ to a fully flexible $[1,256]$ already yields noticeable improvements, as it allows the model to adapt sequence length to input complexity.  
However, very short sequences may be suboptimal for some small images. Setting the minimum limit of tokens to 64 (an example image size of 32×32) yields noticeable gains. Moreover, a larger maximum token limit can further improve text recognition accuracy. Nonetheless, to keep token consumption comparable with other methods, we adopt the range [64, 256] as our baseline.

\begin{figure}[ht]
 \centering
 \includegraphics[width=\linewidth]{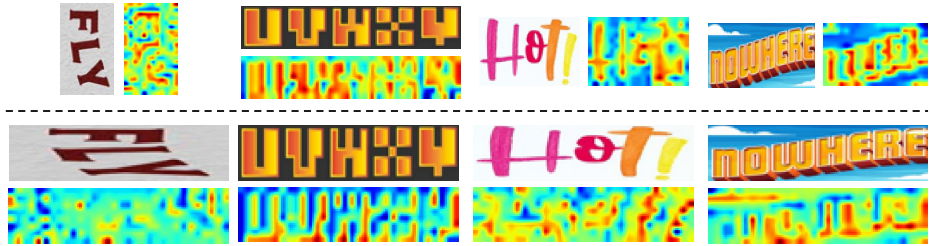}
 \caption{Comparison of feature maps from different encoder outputs. \textbf{Top:} Images fed into the WATERec while preserving aspect ratio, and the corresponding encoder feature maps. \textbf{Bottom:} Baseline with fixed-template inputs (model in the first row of Tab.~\ref{abl}).}
 \label{fig:feature_map}
\end{figure}

\subsection{More Visualization and Analysis}

\subsubsection{Arbitrary-shape modeling directs more robust attention.} To intuitively verify the effectiveness of our arbitrary-shaped input design and to examine whether WATERec indeed robustly handles images with diverse shapes (especially artistic text), we visualize the feature maps produced by the WATERec encoder and by the vanilla ViT baseline with fixed-template inputs, as shown in Fig.~\ref{fig:feature_map}. Benefiting from the preservation of the native aspect ratio, the WATERec encoder accurately and robustly extracts features from artistic text images with diverse shapes and layouts. In contrast, the vanilla ViT baseline stretches vertically oriented text, making the features less salient. This visualization further underscores the importance of supporting arbitrary-shaped inputs for STR (especially in WordArt).

\begin{figure}[ht]
 \centering
 \includegraphics[width=\linewidth]{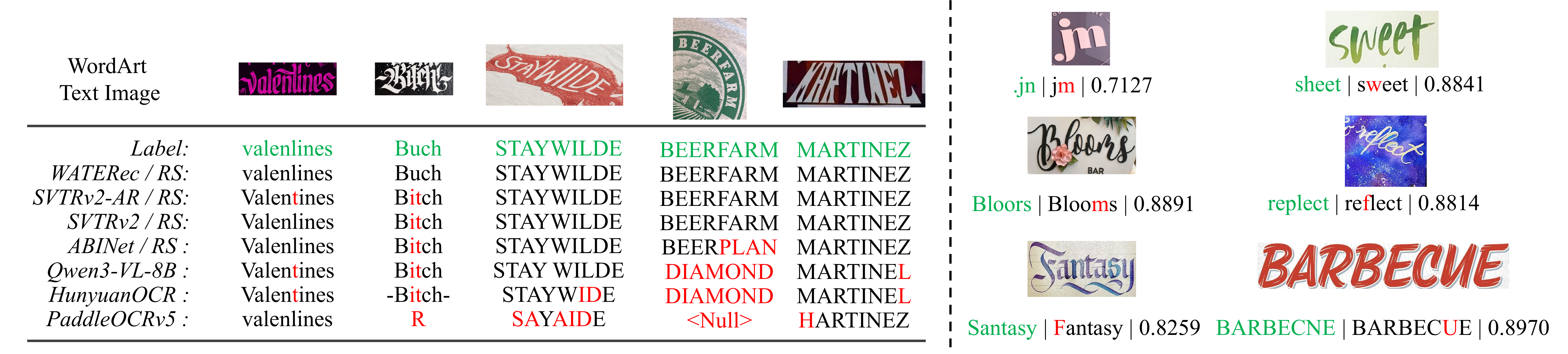}
 \caption{\textbf{Left:} Visualization of different models' predictions on the WordArt-Bench, a supplement to Fig.~\ref{fig:overview}. \textbf{Right:} Some bad-case examples from our final model (WATERec / RS), and the information below each image is formatted as: label | prediction | confidence.
 Characters differing from the ground truth are highlighted in red, and <Null> indicates that there is no output result.}
 \label{fig:result_vis}
\end{figure}

\subsubsection{Our strong baseline still has limitations.} From the left of Fig.~\ref{fig:result_vis}, we observe that WATERec consistently produces predictions matching the ground-truth, even under challenging conditions such as ambiguous characters, distorted fonts, multi-orientations, curved text, and perspective distortions. In contrast, existing STR models are more susceptible to these challenging scenarios and fail on some cases. This type of fine-grained confusion is even more pronounced for VLMs with small but critical errors that severely degrade the overall recognition quality. The right of Fig.~\ref{fig:result_vis} further presents several bad cases of WATERec. Most of these errors arise from inherent ambiguities from a semantic perspective, while some predictions can even be considered reasonable (e.g., ``Blooms'', ``BARBECUE''). We provide the complete set of bad-cases in the appendix, which offers a more comprehensive view of the current limitations of our approach.

\section{Conclusion}

We revisited artistic text recognition from both a data and model perspective, and showed that dedicated design is crucial for handling highly stylized, layout-rich WordArt scenarios. On the data side, we constructed \textbf{WATER-S}, a large-scale synthetic suite composed of one subset rendered with our SynthWordArt engine and the other subset from generative models, together with a carefully deduplicated real training set \textbf{WATER-R}. These datasets complement each other in font controllability, layout diversity, and visual realism, and consistently improve a wide range of STR models. On the model side, we proposed \textbf{WATERec}, which combines a NaViT-like encoder with RoPE and an autoregressive decoder. This design preserves native aspect ratios, mitigates distortion from fixed-template resizing, and better adapts to complex reading orders and artistic layouts. Experiments on WordArt-Bench, common STR benchmarks, and Union14M-Benchmark show that WATERec sets new advanced results, surpasses 90\% accuracy on WordArt-Bench for the first time, and outperforms both general and OCR-specialized VLMs by a clear margin. Our results indicate that artistic text is still a challenging regime for current OCR and VLM systems, and that targeted synthetic data and arbitrary-shape modeling are effective tools for closing this gap. In future, we plan to extend our WATER pipeline to multilingual settings and to support more scripts in both benchmarks and synthetic data. We are also interested in improving VLMs for artistic text recognition through reasoning (e.g., chain-of-thought) to approach expert-model performance.

\section*{Acknowledgements}
This work was supported by the National Natural Science Foundation of China under Grants 625B2057 and 32341012.

\bibliographystyle{splncs04}
\bibliography{main}

\appendix
\clearpage
\setcounter{page}{1}

\title{Appendix}
\author{}
\institute{}

\maketitle

\begin{figure}[ht]
 \centering
 \includegraphics[width=0.6\linewidth]{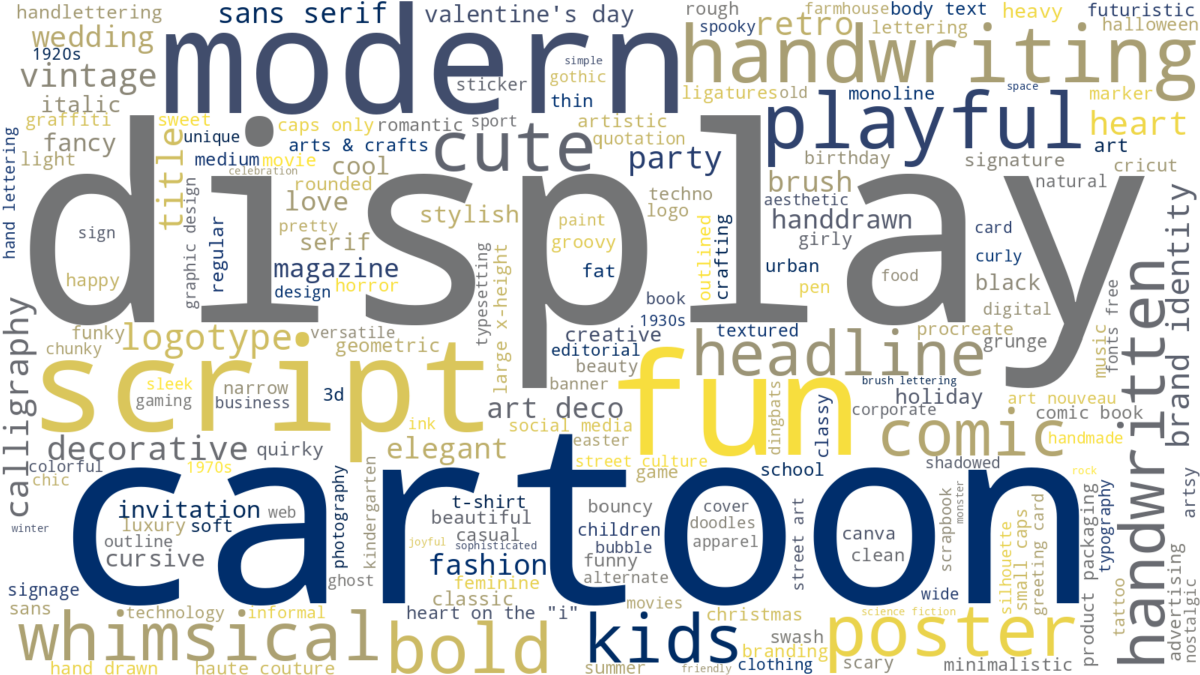}
 \caption{Word cloud of artistic font tags used in WATER-T.}
 \label{fig:font_tags_wordcloud}
\end{figure}

\section{More Data Details}

\subsection{Resources Used and Generated in WATER-S}

\subsubsection{Artistic Fonts}
We first collect artistic font resources from open-source font platforms, design asset websites, and public code repositories. To ensure both stylistic diversity and licensing compliance, we only retain fonts that satisfy the following conditions: 
(1) The font description or tags explicitly mark it as one of the artistic styles (Fig.~\ref{fig:font_tags_wordcloud} shows the distribution of tags), e.g., \texttt{art / display / handwriting / cartoon / playful} or similar. (2) The license allows redistribution or derivative works for research purposes (e.g., OFL, Apache, and some Creative Commons licenses). (3) The font supports complete encodings and normal rendering for English letters and numbers, with distinct glyphs for upper and lower case, so that we can render text accurately and keep the labels clean.
After unifying formats and removing duplicates, we obtain a curated font library containing 11,250 artistic fonts. 

\begin{figure}[ht]
 \centering
 \includegraphics[width=0.9\linewidth]{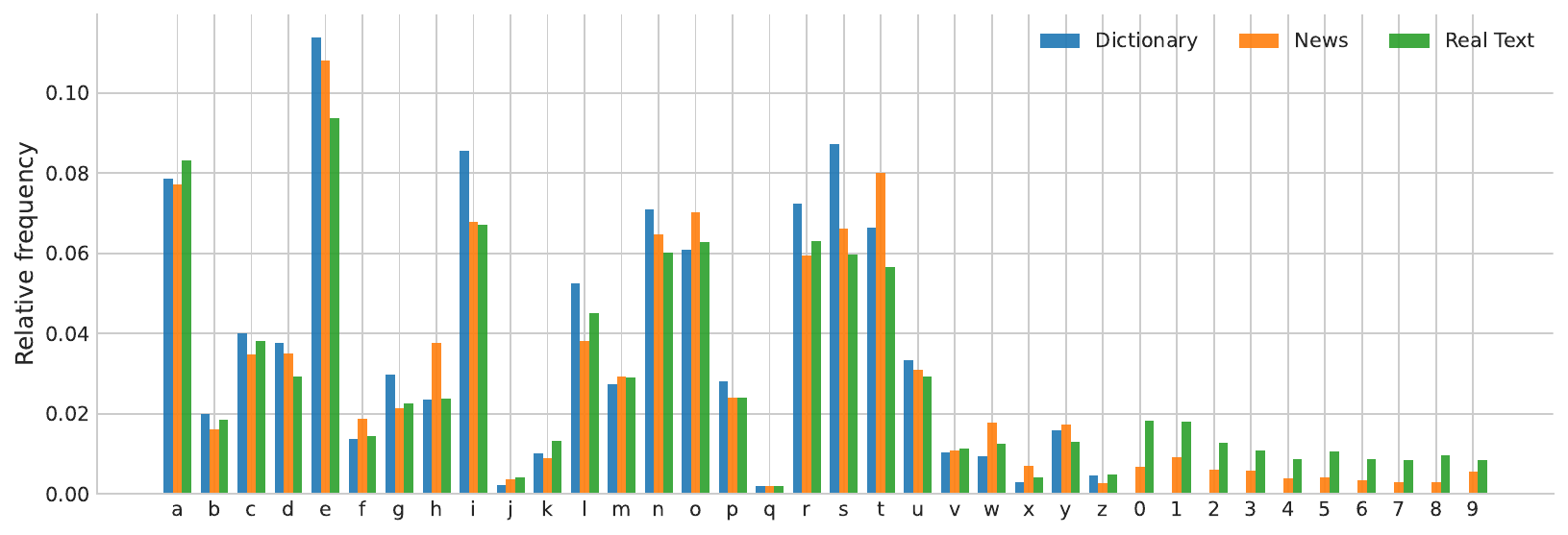}
 \caption{Distribution of representative character proportions across different corpora, with all letters converted to lowercase.}
 \label{fig:char_frequency_comparison}
\end{figure}

\subsubsection{Real Text Corpus}
Conventional synthetic datasets typically sample text labels from the dictionary or news corpora, which further increases the distribution gap (in Fig.~\ref{fig:char_frequency_comparison}) from the real. Here, we directly reuse the text labels from existing large-scale real datasets when rendering our synthetic data. These real texts have 598,615 unique entries, and their lengths are mostly between 1 and 25. 

\subsubsection{Caption Mining Prompt} We use Qwen3-VL-8B~\cite{Qwen3-VL} to caption existing artistic text images. The caption extraction prompt template is as follows:
\begin{AIbox}{Caption Extraction Prompt:}
``I am providing an image of an artistic text area that has been cropped from a real photograph. ``\\
f``The original text within the image is ``\{text\_label\}``.''\\
``Please generate a prompt template that can be used to create images in the same style as this artistic text area. In your prompt, replace the specific text content with \textless Text\textgreater{} so that I can easily substitute different text later. ''\\
``**Requirements:**''\\
``- The prompt should be descriptive, suitable for use with large models (such as generative image models), and accurately capture the artistic and photographic style of the given text image. ''\\
``- Use \textless Text\textgreater{} as a placeholder for the text content within the prompt. ''\\
``- Output only the prompt template. ''\\
``- Do not include the original image, but focus on enabling the generation of a similar style. ''
\end{AIbox}

This procedure produces 31,335 image--caption pairs, one caption per image. An example caption is:
\begin{AIbox}{Caption Example:}
``A close-up, high-resolution photograph of a vintage automotive emblem featuring the text `\textless Text\textgreater{}' rendered in elegant, cursive script with a metallic, brushed silver finish. The lettering is set against a dark, glossy background with subtle reflections and ambient lighting that enhances its three-dimensional appearance. The emblem is mounted on a polished chrome or metallic surface, with visible texture and depth, capturing the classic, luxurious aesthetic of early 20th-century car branding. The composition emphasizes the typography's fluid curves and refined craftsmanship, evoking nostalgia and prestige.''
\end{AIbox}

Next, we perform few-shot mining with Qwen3-VL-8B to obtain additional prompt templates. The few-shot mining prompt is:
\begin{AIbox}{Few-Shot Mining Prompt:}
``You are a professional prompt engineer who creates images of artistic text areas cropped from real photographs. ''\\
``Your task is to generate a new prompt that fully mimics the style, tone, and structural pattern of the provided reference prompts. ''\\
``Here are three examples:\textbackslash n''\\
f``\{caption\_refs[0]\}\textbackslash n''\\
f``\{caption\_refs[1]\}\textbackslash n''\\
f``\{caption\_refs[2]\}\textbackslash n''\\[2pt]
``**Requirements:**''\\
``- The prompt should be descriptive, suitable for use with large models (such as generative image models), and accurately capture the artistic and photographic style. ''\\
``- Use \textless Text\textgreater{} as a placeholder for the text content within the prompt. ''\\
``- Output only the new prompt template, and keep it as concise as possible. ''
\end{AIbox}

We merge the newly mined prompts with the original captions and perform deduplication, resulting in 273,488 distinct prompt templates for artistic text image generation.

\subsection{Filtering of WATER-Z} 
Due to the inherent limitations of generative models, WATER-Z cannot guarantee perfectly correct text rendering. In prior work on synthesizing scene text data, an additional OCR-based filtering step is often applied. For WordArt, however, this is a double-edged sword for some difficult yet correctly rendered samples may be mistakenly removed. We therefore conduct an additional comparison by filtering WATER-Z using our recognition baseline (WATERec / R): we keep only the samples whose predictions exactly match the ground-truth labels, and the measured error rate of WATER-Z under this criterion is 12.56\%. We then re-generate to obtain a filtered dataset of the same scale for a fair comparison. As shown in Tab.~\ref{tab:zdata_f}, the results indicate no clear benefit from filtering and performance on some subsets even fluctuates, suggesting that models can still learn robustly from moderately noisy data. Furthermore, WATER-T can complement WATER-Z in terms of label accuracy, thereby avoiding the influence of certain incorrect results of WATER-Z.

\begin{table*}[ht]
  \centering
  \resizebox{\textwidth}{!}{%
  \setlength\tabcolsep{1pt}
\begin{tabular}{c|c|c|ccccccc|cccccccc}
\toprule
\multirow{2}{*}{\textbf{Training Data}} & \multirow{2}{*}{\textbf{Volume}} & \textbf{WordArt} & \multicolumn{7}{c|}{\textbf{Common Benchmarks}}                                                                & \multicolumn{8}{c}{\textbf{Union14M-Benchmark}}                                             \\
&  & \textbf{Benchmark} &
\textit{IIIT} & \textit{SVT}  & \textit{IC13} & \textit{IC15} & \textit{SVTP} & \textit{CUTE} & \textit{AVG} &
\textit{CUR} & \textit{MLO} & \textit{ART} & \textit{CTL} & \textit{SAL} & \textit{MLW} & \textit{GEN} & \textit{AVG}                  
\\
\midrule

WATER-R + WATER-Z  & 3.2M + 0.5M  & 88.91 & \textbf{99.07} & \textbf{98.30} & \textbf{98.95} & 90.56 & \textbf{95.66} & \textbf{97.92} & \textbf{96.74} & \textbf{93.65} & \textbf{95.11} & \textbf{81.56} & 86.39 & \textbf{89.70} & 89.20 & \textbf{85.25} & \textbf{88.69} \\
WATER-R + WATER-Z-Filter  & 3.2M + 0.5M  & \textbf{89.01} & 99.03&98.15&98.83&\textbf{90.89}&95.50&\textbf{97.92}&96.72&93.12&94.52&81.22&\textbf{86.91}&89.45&\textbf{89.32}&84.98&88.50 \\
\bottomrule
\end{tabular}
  }
\caption{The impact of applying correctness filtering on the training quality of WATER-Z}
\label{tab:zdata_f}
\end{table*}

\subsection{Extending the Image Generator}
The diminishing return at larger synthetic-data scales (Sec.~5.3 of the main paper) is partly related to the distributional upper bound and noise characteristics of a single generative model. In our pipeline, such noise is further mitigated by incorporating the label-accurate WATER-T. Our current experiments already demonstrate the effectiveness of Z-Image-based synthesis. For greater variability, the prompt library is generator-agnostic, so we plan to employ an ensemble of pretrained generative models, including FLUX~\cite{flux}, Qwen-Image~\cite{wu2025qwen}, and the closed-source GPT-Image-2, to further enrich style coverage and reduce model-specific artifacts.

\subsection{Runtime Efficiency of WATER-S} 
To generate WATER-T, we use a server equipped with an Intel Xeon Platinum 8255C CPU (96 cores, 375~GB RAM). With 16 worker processes, generating 1 million samples takes 43,036 seconds, which corresponds to a throughput of approximately 23.24 samples per second. For WATER-Z, we use 8 NVIDIA V100 GPUs with a batch size of 8. Generating 1 million samples takes 10,272 minutes, and each GPU produces on average 730 images per hour. This comparison shows that model-based synthesis, as in WATER-Z, incurs substantially higher computational cost than tool-based synthesis, as in WATER-T. Nevertheless, Z-Image~\cite{cai2025z} is already an efficient image generation model and remains a cost-effective and scalable choice compared with larger closed-source or open-source generative models.

\begin{figure}[ht] \centering
  \includegraphics[width=0.8\textwidth]{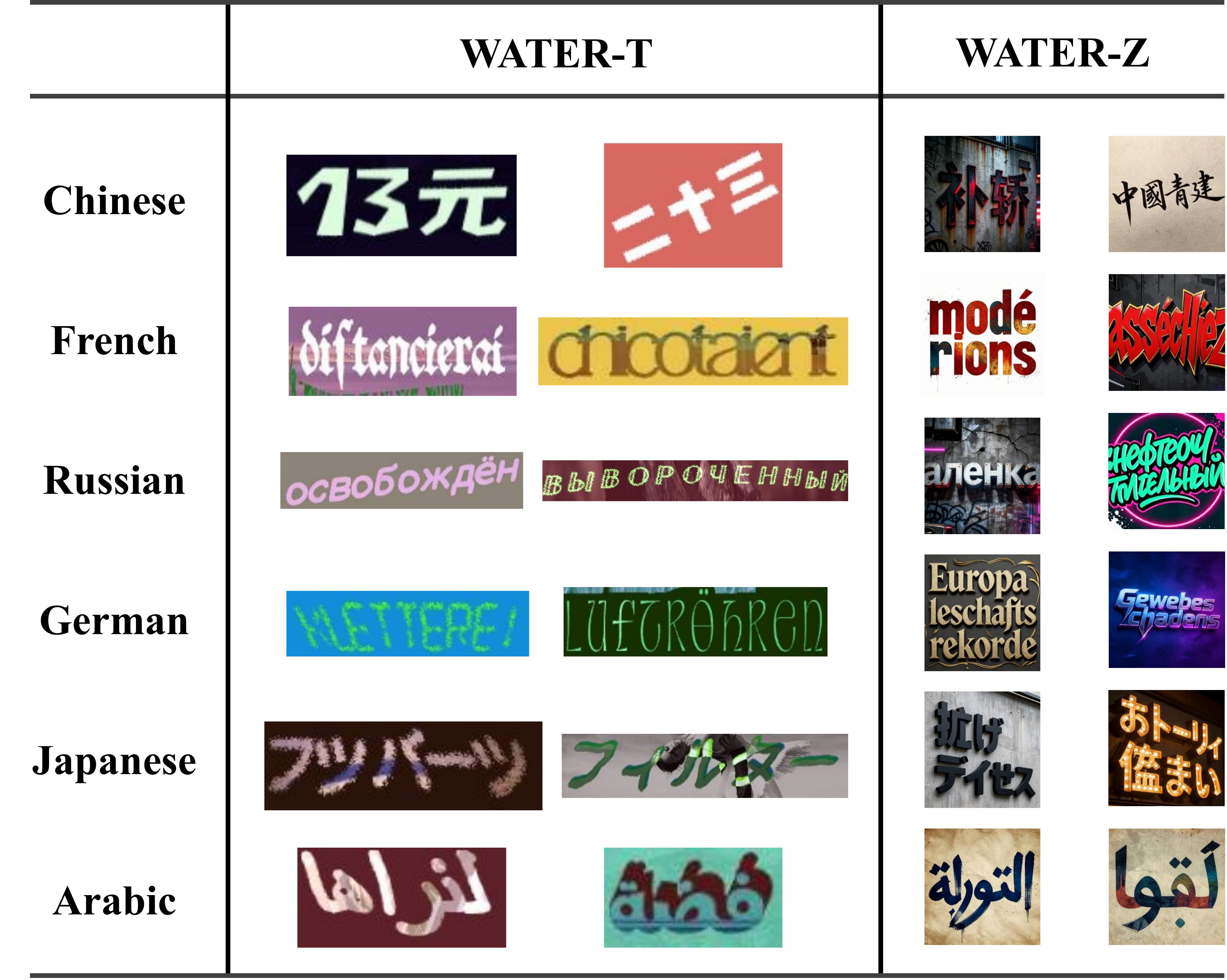}
    \caption{Visualization of synthesized multilingual examples.} \label{fig:mlt}
\end{figure}

\subsection{Multilingual Support}
Both subsets of WATER-S support multiple languages, and we extend them easily by replacing the corresponding language corpus. As shown in Fig.~\ref{fig:mlt}, multilingual examples preserve the same visual style as the English ones.

\begin{table}[ht]
\centering
\footnotesize
\setlength{\tabcolsep}{6pt}
\begin{tabular}{l|c}
\toprule
\textbf{Method / Training Data} & \textbf{Accuracy} \\
\midrule
Qwen3-VL-8B (zero-shot) & 82.77 \\
WATERec / real Chinese data (BCTR-Train) & 87.13 \\
WATERec / real Chinese data + Chinese WATER-S & \textbf{92.08} \\
\bottomrule
\end{tabular}
\caption{Small-scale Chinese artistic-text validation on 101 BCTR-Test WordArt samples.}
\label{tab:chinese}
\end{table}

\subsubsection{Chinese WordArt validation.}
\begin{figure}[ht]
 \centering
 \includegraphics[width=0.9\linewidth]{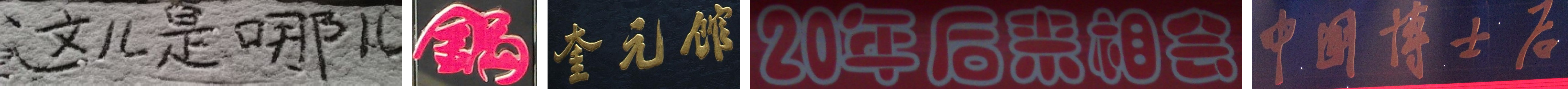}
 \caption{Visualization of the Chinese WordArt test set selected from BCTR-Test~\cite{bctr}.}
 \label{fig:ch_test}
\end{figure}

To more directly support the language-agnostic claim, we conduct an extended validation on Chinese artistic text. We synthesize a 1M-scale Chinese WATER-S by replacing the English corpus with a Chinese one, and select 101 Chinese WordArt samples from BCTR-Test~\cite{bctr} as a small evaluation set (visualized in Fig.~\ref{fig:ch_test}). As reported in Tab.~\ref{tab:chinese}, training WATERec with only real Chinese scene-text data (BCTR-Train) reaches $87.13\%$ accuracy, while adding our synthesized Chinese WATER-S improves it to $92.08\%$. For comparison, the strong general VLM Qwen3-VL-8B~\cite{Qwen3-VL} obtains only $82.77\%$ on the same set. These results confirm that both our synthesis pipeline and the WATERec design transfer effectively to non-Latin scripts.

\subsection{Visualization of WATER-S}
\begin{figure}[ht]
 \centering
 \includegraphics[width=0.9\linewidth]{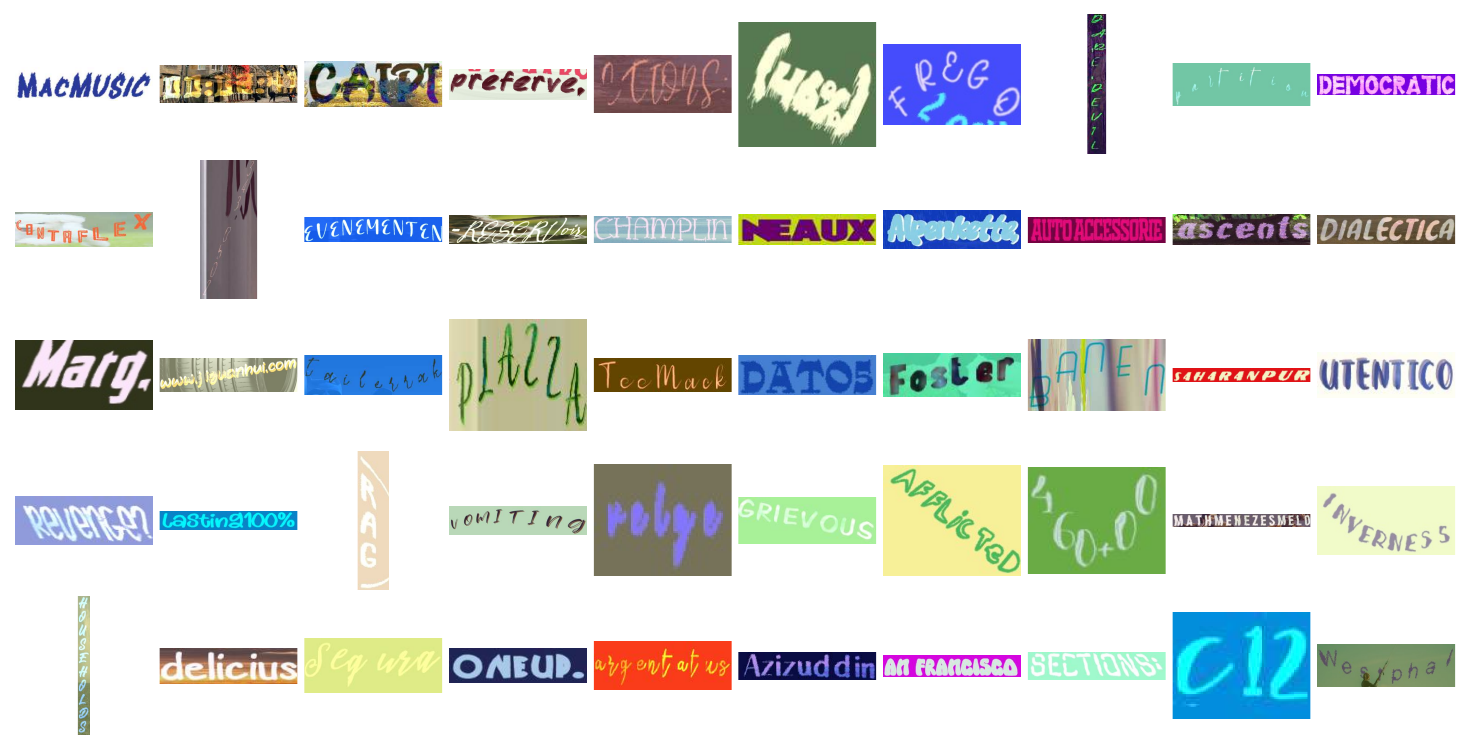}
 \caption{More visualizations of the WATER-T dataset.}
 \label{fig:water_t_more}
\end{figure}

\begin{figure}[ht]
 \centering
 \includegraphics[width=0.9\linewidth]{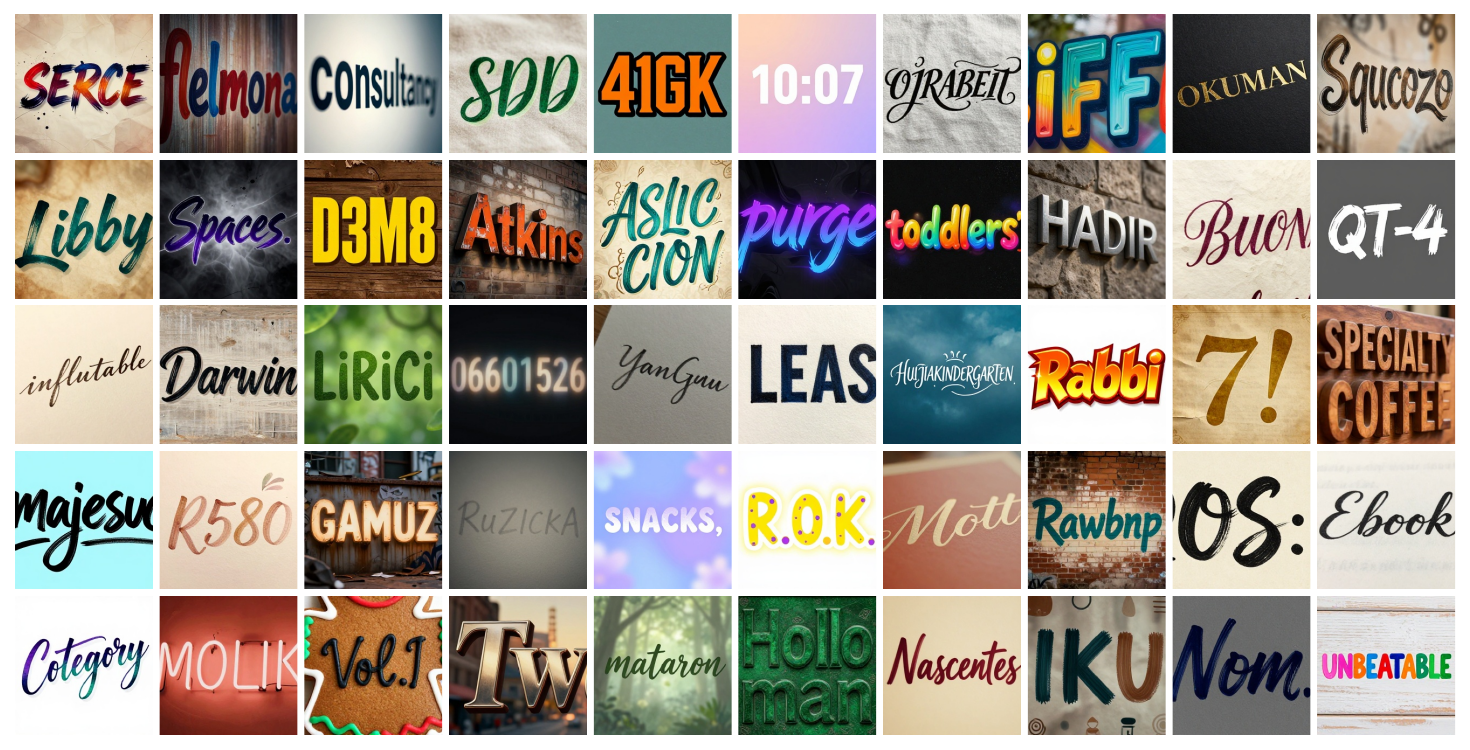}
 \caption{More visualizations of the WATER-Z dataset.}
 \label{fig:water_z_more}
\end{figure}

Fig.~\ref{fig:water_t_more} and Fig.~\ref{fig:water_z_more} respectively show the two subsets of the WATER-S dataset. The WATER-T subset benefits from the customizability of the synthesis engine, so it produces a wider variety of layouts. In contrast, the WATER-Z subset generates styles that better match the aesthetics of human designers, but its layouts, fonts, and other elements are less precisely controllable. This complementary nature makes WATER-S an effective synthetic dataset for WordArt.

\section{More Model Details}

\subsection{Position Embedding}

\subsubsection{APE Implementation}  
Each image is first partitioned into non-overlapping patches and flattened into a token sequence \( \mathbf{x}_n \in \mathbb{R}^{d} \). For each patch we also compute its 2D grid coordinate \( \mathbf{p}_n=(p_n^x,p_n^y) \). We parameterize absolute position as two learnable lookup tables, one for the height axis and one for the width axis. \( \mathbf{E}^{x}\in\mathbb{R}^{L\times d} \) and \( \mathbf{E}^{y}\in\mathbb{R}^{L\times d} \), where \(L\) is the max token length. Given indices \(p_n^x\) and \(p_n^y\), the positional vector is obtained by table indexing and summation,
\[
\mathbf{e}_n = \mathbf{E}^{x}[p_n^x] + \mathbf{E}^{y}[p_n^y]\in\mathbb{R}^{d},
\]
and is added to the patch embedding:
\[
\mathbf{x}'_n = \mathbf{x}_n + \mathbf{e}_n .
\]

\subsubsection{SPE Implementation}  
We construct a deterministic 2D sine-cosine embedding \(\mathbf{e}_n\in\mathbb{R}^{d}\) from the patch coordinate \(\mathbf{p}_n=(p_n^x,p_n^y)\). Assume \(d\) is divisible by \(4\), and define frequencies
\[
\omega_t = 10000^{-\,t/(d/4)},\qquad t=0,\dots,\frac{d}{4}-1.
\]
We allocate half of the channels to the horizontal axis and half to the vertical axis, and define
\[
\mathbf{e}^x_n=\big[\sin(\omega_0 p_n^x),\cos(\omega_0 p_n^x),\dots,\sin(\omega_{d/4-1} p_n^x),\cos(\omega_{d/4-1} p_n^x)\big]\in\mathbb{R}^{d/2},
\]
\[
\mathbf{e}^y_n=\big[\sin(\omega_0 p_n^y),\cos(\omega_0 p_n^y),\dots,\sin(\omega_{d/4-1} p_n^y),\cos(\omega_{d/4-1} p_n^y)\big]\in\mathbb{R}^{d/2}.
\]
The final 2D SPE is the concatenation
\[
\mathbf{e}_n = [\mathbf{e}^x_n;\mathbf{e}^y_n]\in\mathbb{R}^{d},
\]
and it is injected into the token representation by additive composition:
\[
\mathbf{x}'_n=\mathbf{x}_n+\mathbf{e}_n.
\]

\subsubsection{RoPE Implementation}
Formally, we compute query and key vectors
$ \mathbf{q}_n, \mathbf{k}_n \in \mathbb{R}^{d} $ from the input tokens and convert them into complex form by pairing every two consecutive channels as the real and imaginary parts: $
\bar{\mathbf{q}}_n, \bar{\mathbf{k}}_n \in \mathbb{C}^{d/2} $.
We then follow the axial formulation and allocate half of
the complex channels to the horizontal axis and half to the vertical axis.
Concretely, we define frequency coefficients for the two axes as
\[
\omega_t^x = \theta^{-\,t / d}, \quad
\omega_t^y = \theta^{-\,t / d}, \qquad
t = 0, \dots, \frac{d}{4} - 1,
\]
where \(\theta\) is a base frequency (we use \(\theta = 100\) in our implementation).
For a patch at position \(\mathbf{p}_n\), the corresponding complex rotation is
constructed as
\[
\mathbf{R}_n =
\big[\,e^{i\,\omega_0^x p^x_n}, \dots, e^{i\,\omega_{d/4-1}^x p^x_n},
      e^{i\,\omega_0^y p^y_n}, \dots, e^{i\,\omega_{d/4-1}^y p^y_n}\big]
\in \mathbb{C}^{d/2}.
\]
The rotary embedding is then applied to queries and keys via element-wise
complex multiplication:
\[
\bar{\mathbf{q}}_n' = \bar{\mathbf{q}}_n \circ \mathbf{R}_n, 
\qquad
\bar{\mathbf{k}}_n' = \bar{\mathbf{k}}_n \circ \mathbf{R}_n,
\]
and finally, we convert these two vectors back from their complex form to the real domain: $ \mathbf{q}_n', \mathbf{k}_n' \in \mathbb{R}^{d} $. This allows the model to capture relative relations without explicit learnable positional vectors and naturally supports variable-length sequences, with strong generalization across different scales and resolutions.

\begin{table*}[ht]
\centering
\begin{tabular}{c|c|c|c}
\toprule
\textbf{Type} & \textbf{Method} & \textbf{FPS$\uparrow$} & \textbf{Param.$\downarrow$} \\
\midrule

\multirow{2}{*}{\textbf{CTC}} & CRNN~\cite{shi2017crnn} & 974.26 & 16.20M \\
 & SVTRv2~\cite{du2024svtrv2} & 608.48 & 21.02M \\
 \midrule
 
\multirow{2}{*}{\textbf{PD}} & ABINet~\cite{abinet} & 899.12 & 36.86M \\
 & CPPD~\cite{cppd} & 640.29 & 26.96M \\
 \midrule
 
\multirow{4}{*}{\textbf{AR}} & SVTRv2-AR & 500.42 & 22.46M \\
 & MAERec~\cite{jiang2023revisiting} & 305.78 & 35.69M \\
 & WATERec & 361.66 & 26.22M \\
 & Vanilla ViT Baseline & 364.78 & 26.12M \\

\bottomrule
\end{tabular}

\caption{Comparison of inference speed in frames per second (FPS) and model parameters for different STR methods on the validation set, calculated using the OpenOCR framework on an NVIDIA V100 GPU.}
\label{tab:comp_cost}
\end{table*}

\subsection{Computational Cost of WATERec} 
As shown in Tab.~\ref{tab:comp_cost}, we compare the runtime cost of different types of STR models under the same setting. NAR approaches such as PD and CTC achieve higher efficiency. WATERec, which uses a Transformer-based architecture, runs slightly slower in terms of FPS than hybrid CNN-Transformer models such as SVTRv2, but it still outperforms MAERec, which also adopts a pure Transformer architecture. In addition, compared with the vanilla ViT baseline, WATERec introduces no extra parameters and therefore attains a similar FPS. We further measure the effect of the token range on efficiency: when enlarging the range from $[64,256]$ to $[64,512]$, the FPS of WATERec drops from 361.66 to 191.46. Although a larger token budget can slightly improve accuracy (see Tab.~\ref{abl} in the main paper), $[64,256]$ provides a clearly better accuracy/efficiency trade-off, which is why we adopt it as the default.

\subsection{Analysis of the C-Bench Trade-off}
On the regular C-Bench, the fixed-resolution MAERec~\cite{jiang2023revisiting} slightly outperforms WATERec (Tab.~1 in the main paper). We attribute this to the distributional difference between regular scene text and WordArt: regular text has a relatively stable and near-horizontal aspect-ratio distribution, for which a ViT backbone with fixed-size resizing is close-to-optimal and aspect-ratio preservation matters less. In contrast, WordArt exhibits highly variable shapes, layouts, and extreme aspect ratios, where preserving the native aspect ratio is essential and brings clear gains on A-Bench and the harder U-Bench subsets (e.g., \textit{ART} and \textit{SAL}). Hence, the small drop on C-Bench (within $0.3\%$ of the best) is an expected and acceptable trade-off in exchange for substantially stronger WordArt recognition, and WATERec still remains highly competitive on regular text.

\section{More Evaluation Details}

\subsection{Evaluation of VLMs}
Because our evaluation goal is to directly recognize the text in images, we use the following fixed prompt template for general VLMs:
\begin{AIbox}{General Prompt:}
``Please directly output all original readable text in the image.'' \\
``Do not include any explanation or description. Only output the recognized text.''
\end{AIbox}

For OCR-specialized VLMs, we instead adopt their official prompts that are designed to output text only. The prompts for each model are as follows:
\begin{AIbox}{DeepSeek-OCR Prompt:}
``<image>\textbackslash nFree OCR.''
\end{AIbox}
\begin{AIbox}{PaddleOCR-VL Prompt:}
``OCR:''
\end{AIbox}
\begin{AIbox}{GOT-OCR2.0 Prompt:}
``ocr''
\end{AIbox}
\begin{AIbox}{HunyuanOCR Prompt:}
``Extract the text from this image.''
\end{AIbox}

For PP-OCRv5~\cite{cui2025paddleocr}, we use the recognition model \texttt{PP-OCRv5\_server\_rec} and only call this recognition component. All other parameters follow the official default settings without any additional modifications.

\begin{table}[ht]
\centering
\footnotesize
\setlength{\tabcolsep}{4pt}
\begin{tabular}{l|c|c|c}
\toprule
\textbf{Model} & \textbf{\#Params} & \textbf{Training Data} & \textbf{A-Bench} \\
\midrule
Qwen3-VL-8B (zero-shot) & 8B  & --              & 72.01 \\
Qwen3-VL-8B + SFT       & 8B  & WATER-R         & 82.59 \\
Qwen3-VL-8B + SFT       & 8B  & WATER-R+WATER-S & 84.78 \\
\midrule
WATERec (ours)          & $\sim$26M & WATER-R+WATER-S & \textbf{90.40} \\
\bottomrule
\end{tabular}
\caption{Expert baseline vs.\ SFT on a strong general VLM, evaluated on A-Bench.}
\label{tab:vlm_sft}
\end{table}

\subsection{Fine-tuning VLMs with WATER-S}
In real-world OCR applications, compact task-specialized recognizers remain a dominant choice, as they offer clear advantages in deployment cost, resource consumption, and adaptation to minority languages and difficult scenarios. To examine whether a strong general VLM can close the gap on WordArt after adaptation, we fine-tune Qwen3-VL-8B~\cite{Qwen3-VL} with LoRA~\cite{hu2022lora} for 20k steps using the ms-swift framework and our data. As shown in Tab.~\ref{tab:vlm_sft}, supervised fine-tuning (SFT) brings clear gains (from $72.01\%$ zero-shot to $82.59\%$ with WATER-R, and further to $84.78\%$ with WATER-R+WATER-S), and WATER-S again provides a consistent improvement, confirming that our synthetic data is also useful for VLM adaptation. Nevertheless, the fine-tuned 8B VLM still does not surpass our lightweight ($\sim$26M) expert WATERec ($90.40\%$). This suggests that, although WATER-S benefits VLM adaptation, a task-specialized recognizer remains more effective and efficient for difficult WordArt recognition.

\begin{figure}[ht]
 \centering
 \includegraphics[width=0.9\linewidth]{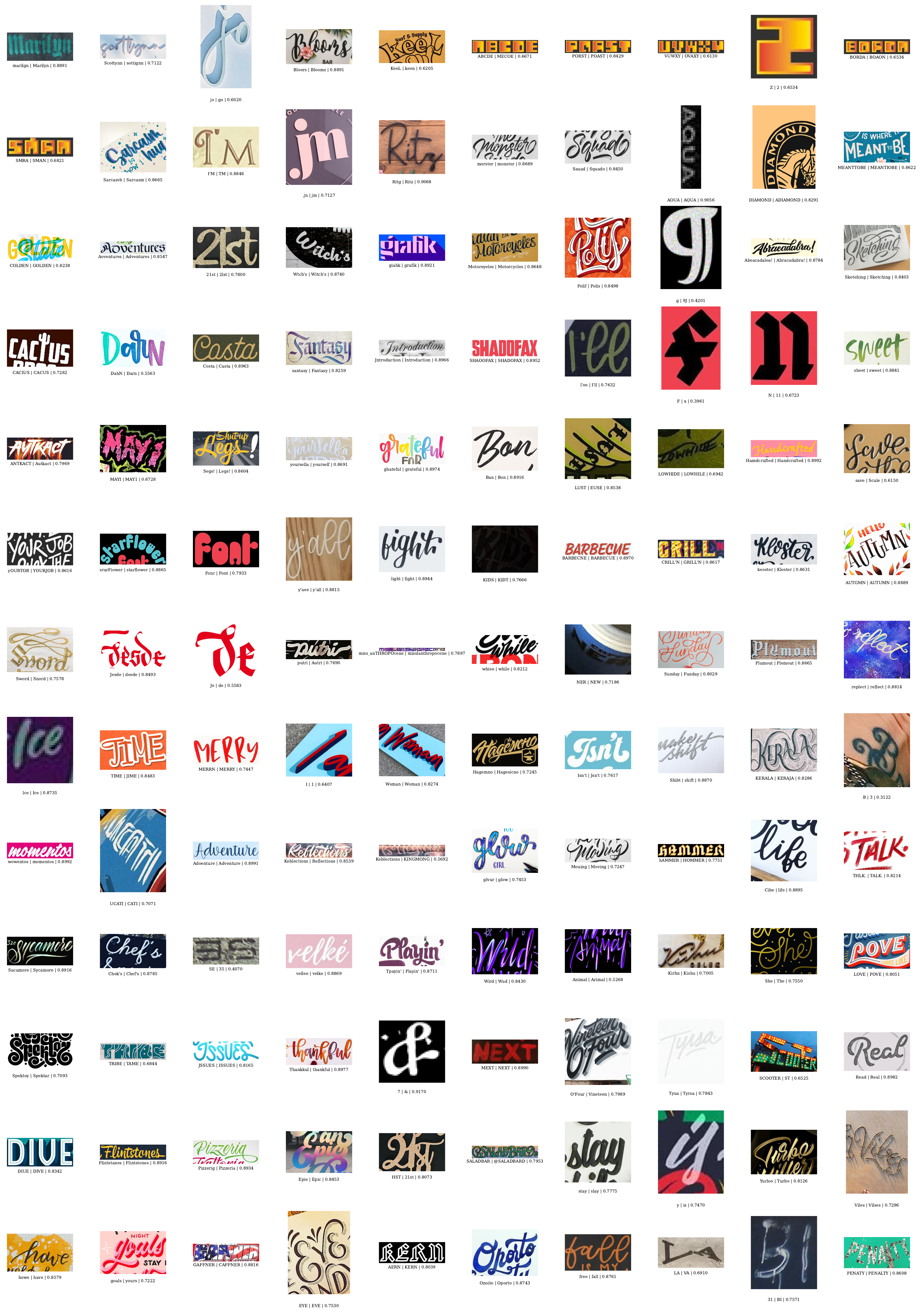}
 \caption{More bad-case examples from our final model (WATERec / RS), and the information below each image is formatted as: label | prediction | confidence.}
 \label{fig:bad_all}
\end{figure}

\end{document}